\documentclass{article}
\usepackage{spconf,amsmath,graphicx,hyperref}

\usepackage{times}
\usepackage{latexsym}

\usepackage[T1]{fontenc}
\usepackage[utf8]{inputenc}

\usepackage{microtype}

\usepackage{inconsolata}

\usepackage{tikz}
\usepackage{amsmath}
\usepackage{amsmath}
\usepackage{amssymb}
\usepackage{mathtools}
\usepackage{amsthm}
\usepackage{multirow}
\usepackage{csquotes}
\usepackage{algorithm}
\usepackage{algorithmic}
\usepackage{epstopdf}
\usepackage{stfloats}
\usepackage{booktabs}
\usepackage{enumitem}
\usepackage{tabularx}
\usepackage{tikz,lipsum}
\usepackage[most]{tcolorbox}
\usepackage{wrapfig}

\usepackage{graphicx}

\title{AutoEval: A Practical Framework for Autonomous Evaluation of Mobile Agents}

\name{Jiahui Sun, Zhichao Hua, Yubin Xia}
\address{Shanghai Jiao Tong University, Shanghai, China\\
Email: \{jason\_2001, zchua, xiayubin\}@sjtu.edu.cn}

\begin{document}
\ninept

% Our project name
\newcommand{\sys}{AutoEval}

\newcommand\TODO[1]{\textcolor{red}{TODO: #1}}
\newcommand\CHECK[1]{\textcolor{blue}{CHECK: #1}}

\newcommand\RED[1]{\textcolor{red}{#1}}
\newcommand\BLUE[1]{\textcolor{blue}{#1}}

\definecolor{main}{HTML}{4472C4}    % setting main color to be used
\definecolor{sub}{HTML}{EBF4FF}     % setting sub color to be used

%-------------------------------------------------------------------------------
\maketitle
%-------------------------------------------------------------------------------

% \noindent\rule{80mm}{2pt}
\begin{abstract}
Comprehensive evaluation of mobile agents can significantly advance their development and real-world applicability.
However, existing benchmarks lack practicality and scalability due to the extensive manual effort in defining task reward signals and implementing evaluation codes.
We propose \sys{}, an evaluation framework which tests mobile agents without any manual effort.
Our approach designs a UI state change representation which can be used to automatically generate task reward signals, and employs a Judge System for autonomous evaluation.
Evaluation shows \sys{} can automatically generate reward signals with high correlation to human-annotated signals, and achieve high accuracy (up to 94\%) in autonomous evaluation comparable to human evaluation.
Finally, we evaluate state-of-the-art mobile agents using our framework, providing insights into their performance and limitations.
\end{abstract}
% %-------------------------------------------------------------------------------
% \begin{abstract}
% %-------------------------------------------------------------------------------
% Your abstract text goes here. Just a few facts. Whet our appetites.
% Not more than 200 words, if possible, and preferably closer to 150.
% \end{abstract}

%-------------------------------------------------------------------------------
\section{Introduction}

\definecolor{myred}{rgb}{0.8,0.2,0.2}
\definecolor{myblue}{HTML}{2e75b6}
\definecolor{myyellow}{HTML}{C2953F}
\definecolor{mypurple}{HTML}{663399}

% Task and application
The advancement of Large Language Models (LLMs) has empowered mobile agents to understand and interact effectively with mobile platforms like Android, potentially freeing humans from tedious daily tasks~\cite{Zhang2023AppAgentMA, zhang-zhang-2024-look, Nong2024MobileFlowAM, Lee2023ExploreSD, Wen2024AutoDroid, Wang2024MobileAgentv2MD,Hong2023CogAgentAV, wang2025mobile, qin2025ui}.
However, these agents still struggle to perform reliably even on simple tasks.
Therefore, benchmarking mobile agent performance is critical. 
An effective benchmark with fine-grained feedback reveals the agent's bottlenecks and guides targeted improvements.

% previous work
However, existing mobile agent benchmarks are not practical enough to evaluate.
Some benchmarks~\cite{li2020mapping, Rawles2023AndroidIW, xing2024androidarena, zhang2024llamatouch,li2024effects, venkatesh2022ugif,klissarov2023motif} compare agent trajectories with human demonstrations, but this approach is misleading since agents can complete tasks through valid paths that differ from the demonstrations.
For this reason, recent benchmarks~\cite{xu2024androidlab, rawles2024androidworld} choose to evaluate agents leveraging reward signals like expected UI state or operating system state.
However, such benchmarks are impractical for two reasons.
First, they still require significant manual effort to define task-specific reward signals.
Second, checking these reward signals requires extensive code development (e.g., 3,000 lines for 138 tasks in AndroidLab~\cite{xu2024androidlab}).
To address the manual effort issue, previous work Agent-Eval-Refine~\cite{Pan2024AutonomousEA} designs a LLM-based evaluator for agent evaluation.
However, it only provides a binary answer of whether the agent completes the task or not, which is too coarse-grained.
Our work builds on top of these advancements, seeking a fine-grained autonomous and reliable evaluation of mobile agents.

\begin{figure*}[t!]
\centering
\includegraphics[width=\textwidth]{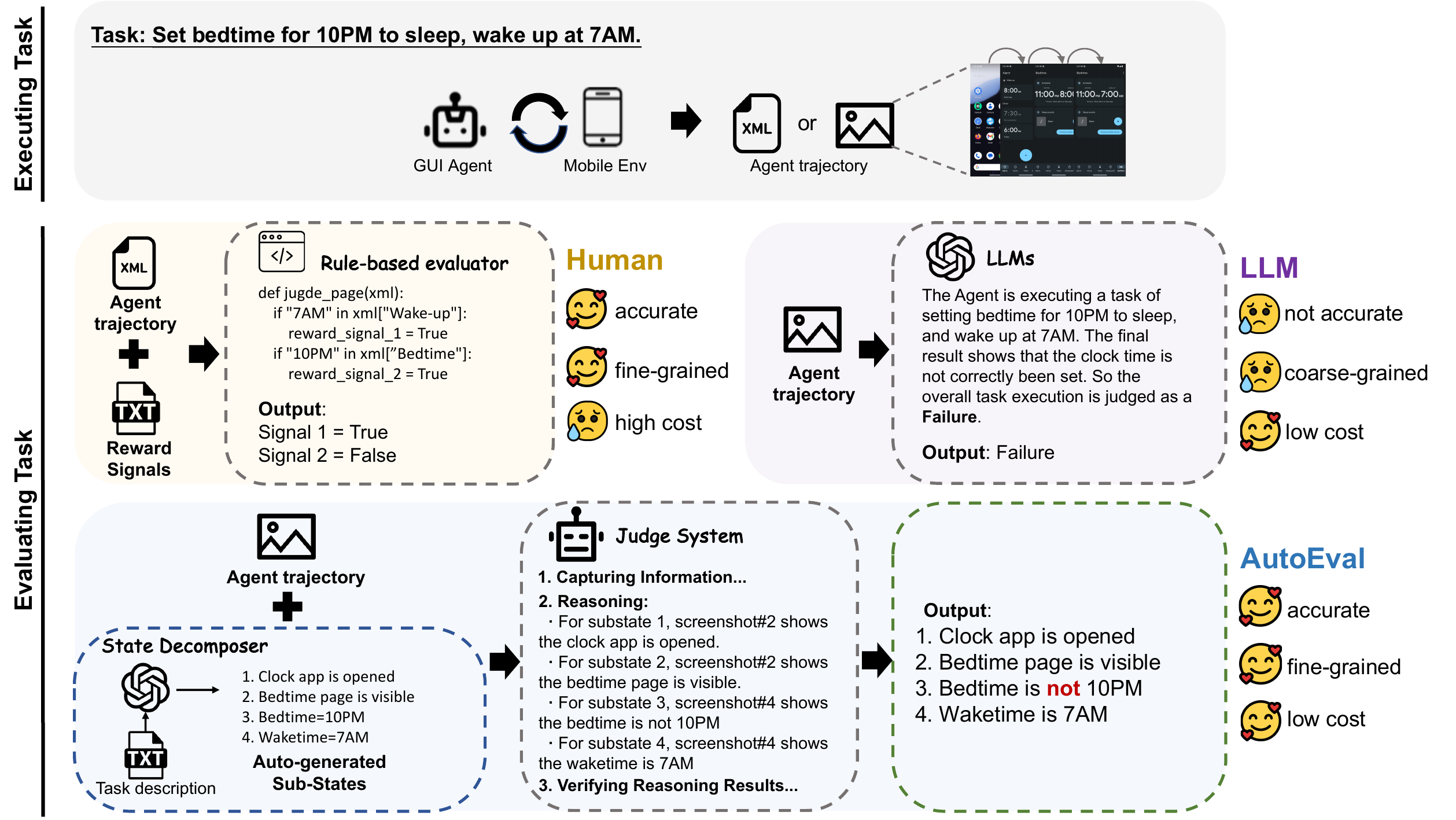}
\caption{Comparison of \textcolor{myblue}{\textbf{AutoEval}} (ours) with existing agent benchmarks. 
\textcolor{myyellow}{\textbf{Human}}-based-benchmark~\cite{xu2024androidlab,rawles2024androidworld}: requires manual reward signal definition and extensive code development for agent evaluation, leading to high accuracy but high cost.
\textcolor{mypurple}{\textbf{LLM}}-based-benchmark~\cite{Pan2024AutonomousEA}: inputs the agent's execution trajectory and utilizes LLM to generate a binary answer of whether the agent completes the task or not. 
It reduces the costs significantly but sacrifices the accuracy and fine-grained agent performance feedback.
\textcolor{myblue}{\textbf{AutoEval}}: introduces automatically generated substates along with an autonomous Judge System to provide fine-grained feedback on agent's performance, achieving both high accuracy and low cost.
}
\label{fig:teaser}
\end{figure*}

In this paper, we propose \sys{}, a novel autonomous evaluation framework for mobile agents.
The basic idea is illustrated in Figure~\ref{fig:teaser}.
% Design Challenge 
% TODO: 保证下面的描述和figure 1是对应的
The key technical challenge in our work lies in how to design a LLM-based pipeline to generate task reward signals and evaluate agent performance autonomously while keeping accuracy simultaneously.
To tackle this challenge, we design a Structured Substate Representation model to describe task reward signal and implement State Decomposer (based on LLMs) which can automatically generate substates accordingly.
Moreover, we develop a Judge System that autonomously evaluates agent performance through a three-stage pipeline:
(1) capturing textual description from agent execution screenshots
(2) evaluating agent performance by reasoning about screenshot descriptions against generated substates
(3) checking evaluation results through consistency constraints to ensure the reliability of the evaluation results.
Based on the above design, {\sys} can evaluate mobile agents without any manual effort.

We collect tasks from existing mobile agent benchmarks and augment them with automatically generated substates.
Comparing these generated substates with human-annotated ones, we find that they cover 93.28\% of human-annotated substates without additional knowledge.
We then evaluate our Judge System's reliability, achieving 94.35\% accuracy compared with oracle human evaluation.
Finally, we conduct comprehensive evaluation of state-of-the-art mobile agents using our benchmark, providing detailed analysis of their performance characteristics and failure patterns.

\textbf{Our Contributions can be summarized as follows:}
\begin{itemize}
\setlength{\itemsep}{0pt}
\setlength{\parskip}{0pt}
\setlength{\parsep}{0pt}
    \item We propose Structured Substate Representation with State Decomposer for autonomous substate generation, and create a benchmark with 93 tasks from existing datasets.
    \item We design an autonomous Judge System that replaces human-written evaluation code and provides fine-grained and reliable performance feedback.
    \item We implement a prototype system that implements the proposed design and evaluate the effectiveness of our method.
    \item We test existing mobile agents in our framework, comparing and analyzing their performance in a fine-grained manner.
\end{itemize}

%-------------------------------------------------------------------------------
\section{Related Work}
\subsection{GUI Agents}
Before the prevalence of Large Language Model, 
traditional autonomous agents primarily implement through reinforcement learning \cite{Branavan2009ReinforcementLF, Shvo2021AppBuddyLT, Gur2022EnvironmentGF},
semantic parsing \cite{li-etal-2020-mapping} and imitation learning~\cite{Humphreys2022ADA} that clones human's keyboard and mouse actions .

The recent trend is to use Large Language Model to generate GUI instructions and actions. 
A series of works leverage prompt engineering to accomplish automation tasks through both text-based~\cite{Lee2023ExploreSD, Wen2024AutoDroid,wang2025mobile, Taeb2023AXNavRA} and multi-modal prompts \cite{Zhang2023AppAgentMA, Wang2024MobileAgentv2MD, Lu2024OmniParserFP}.
Another line of work has concentrated on improving the performance of GUI agents through LLM training, 
utilizing task-specific model architecture \cite{zhang-zhang-2024-look,Nong2024MobileFlowAM, Hong2023CogAgentAV},
supervised fine-tuning \cite{Lu2024GUIOA}, and reinforcement learning \cite{Bai2024DigiRLTI}.
More recent agents \cite{Zhang2023AppAgentMA,Pan2024AutonomousEA, Li2023AZL, Wu2024OSCopilotTG} explores using inference-time techniques like reflection \cite{Shinn2023ReflexionLA}, self-critique, online-exploration to improve the agent performance from trial and error.

\subsection{Mobile Agent Benchmark} 
Recent benchmarks for evaluating mobile agents can be categorized into two types based on their evaluation metrics: reference-based benchmarks and reward-signal-based ones.

Reference-based benchmark normally compares agent's action trajectory with human reference trajectory, calculating an action matching score by measuring the similarity between the two trajectories.
This type of benchmark requires burdensome human effort to collect reference trajectories, making it expensive and time-consuming to scale. For example, AitW \cite{Rawles2023AndroidIW} collects over 715K reference trajectories, highlighting the significant resource investment needed for reference-based evaluation.
Moreover, although some benchmarks \cite{xing2024androidarena, zhang2024llamatouch} propose different kinds of action matching functions to enhance evaluation accuracy, 
reference-based benchmarks cannot truly evaluate agent performance in real-world environments,
as agents may successfully complete tasks through valid alternative paths that differ from the human reference trajectories.

In contrast, reward-signal-based benchmarks evaluate agent performance by checking environment reward signals for task completion, 
providing a more reliable and realistic evaluation metrics.
For example, AndroidWorld \cite{rawles2024androidworld} leverages the state management capabilities of the Android operating system like file system state as task reward signals. 
Other benchmarks like \cite{xu2024androidlab} pull XML description of the UI during agent execution and check specific UI state changes like text content and button availability as reward signals.
However, these benchmarks still require human effort to define the reward signals, like inspecting how file system state changes while agent performing application tasks\cite{rawles2024androidworld}. 
Additionally, checking reward signal often involves tedious coding work, such as XML string pattern matching\cite{xu2024androidlab}.
Our work takes inspiration from reward-signal-based benchmarks and aims to develop a reliable and autonomous approach to define and check task reward signals.

\subsection{Autonomous Evaluation}
Several works~\cite{Zhuge2024AgentasaJudgeEA,fu-etal-2024-gptscore, Chen2024MLLMasaJudgeAM, Zhao2024AutoArenaAL, Chan2023ChatEvalTB} 
have explored using LM-based system to achieve automated and scalable evaluation. 
For example, the work \cite{Zhuge2024AgentasaJudgeEA} propose Agent-as-a-Judge, designing a sophisticated agentic system that can autonomously evaluate the performance of coding agents,
outperforming previous LLM-as-a-Judge \cite{Zheng2023JudgingLLM} and demonstrates reliability comparable to human evaluation.
% 在mobile environment 场景的auto evaluation 比较少见 
Autonomous evaluation for mobile agents remains relatively unexplored. 
Previous work Agent-Eval-Refine \cite{Pan2024AutonomousEA} designs a model-based evaluator to provide evaluation of a mobile agent's trajectory
However, it provides only a binary answer of whether the agent completes the task or not, which is primarily used as the reward function for Reflexion \cite{Shinn2023ReflexionLA} or filtered behavior cloning.
Our work builds on top of these advancements, seeking a fine-grained autonomous and reliable evaluation of mobile agents.

%-------------------------------------------------------------------------------
\section{System Design}

\subsection{Overview}
\label{subsec:overview}
We propose \sys{}, an autonomous evaluation framework for mobile agents that aims to fulfill two design goals (DG).

\textbf{DG1. Autonomous: Minimize human effort in evaluating mobile agents.} 
\sys{} seeks to automate the entire evaluation process by eliminating manual definition of task reward signals and paired evaluation code.

\textbf{DG2. Reliable: Comparable evaluation accuracy to human evaluation.}

To achieve DG1, \sys{} uses a State Decomposer that autonomously generates a specific task's substates.
Then, Judge System (Section \ref{subsec:Judge System}) autonomously evaluates the agent's performance by checking substates completion given screenshots-trajectory during the agent's task execution.
% TODO: 描述一下为什么要引入Structured Substate Representation Model以及Judge System

To achieve DG2, we design Structured Substate Representation (Section \ref{subsec:substate-model}) that captures UI states as reward signals during task execution.
We guide the Judge System to evaluate each substate using a reasoning-then-checking approach.

\subsection{Structured Substate Representation Model}
\label{subsec:substate-model}
In this section, we introduce the \textbf{S}tructured \textbf{S}ubstate \textbf{R}epresentation Model (SSR) which describes the UI state during the agent's task execution.

% Substates are defined as some unescapable UI states like \enquote{Firefox's main page is visible}, indicating the correctness of agent task execution, serving the same purpose as reward signals.
% Substates can be further represented as StateNodes; each StateNode contains a natural language description of a specific UI state and a pointer to its parent StateNode.
% Based on the observation that the agent's interaction with mobile apps primarily consists of navigating between pages and performing operations (e.g., typing and clicking) within each page, 
% StateNodes are then categorized into two types: PageNodes and UnitNodes.
Substates are UI states that indicate task execution correctness.
They are represented as StateNodes with natural language descriptions and parent pointers, categorized into PageNodes and UnitNodes based on whether they represent page navigation or in-page operations. All StateNode's parent must be another PageNode.
\begin{itemize}
\vspace{-0.4em}
\setlength{\itemsep}{0pt}
\setlength{\parskip}{0pt}
\setlength{\parsep}{0pt}
    \item \textbf{PageNode}: represents page state (e.g., \enquote{The app's search page is visible}). 
    \item \textbf{UnitNode}: represents unit state (e.g., \enquote{The search input field contains \enquote{Large Language Model}}). 
\end{itemize}
\vspace{-0.4em}

\subsection{State Decomposer}
\label{subsec:state-decomposer}
\begin{figure}[htbp]
\centering
\includegraphics[width=0.95\columnwidth]{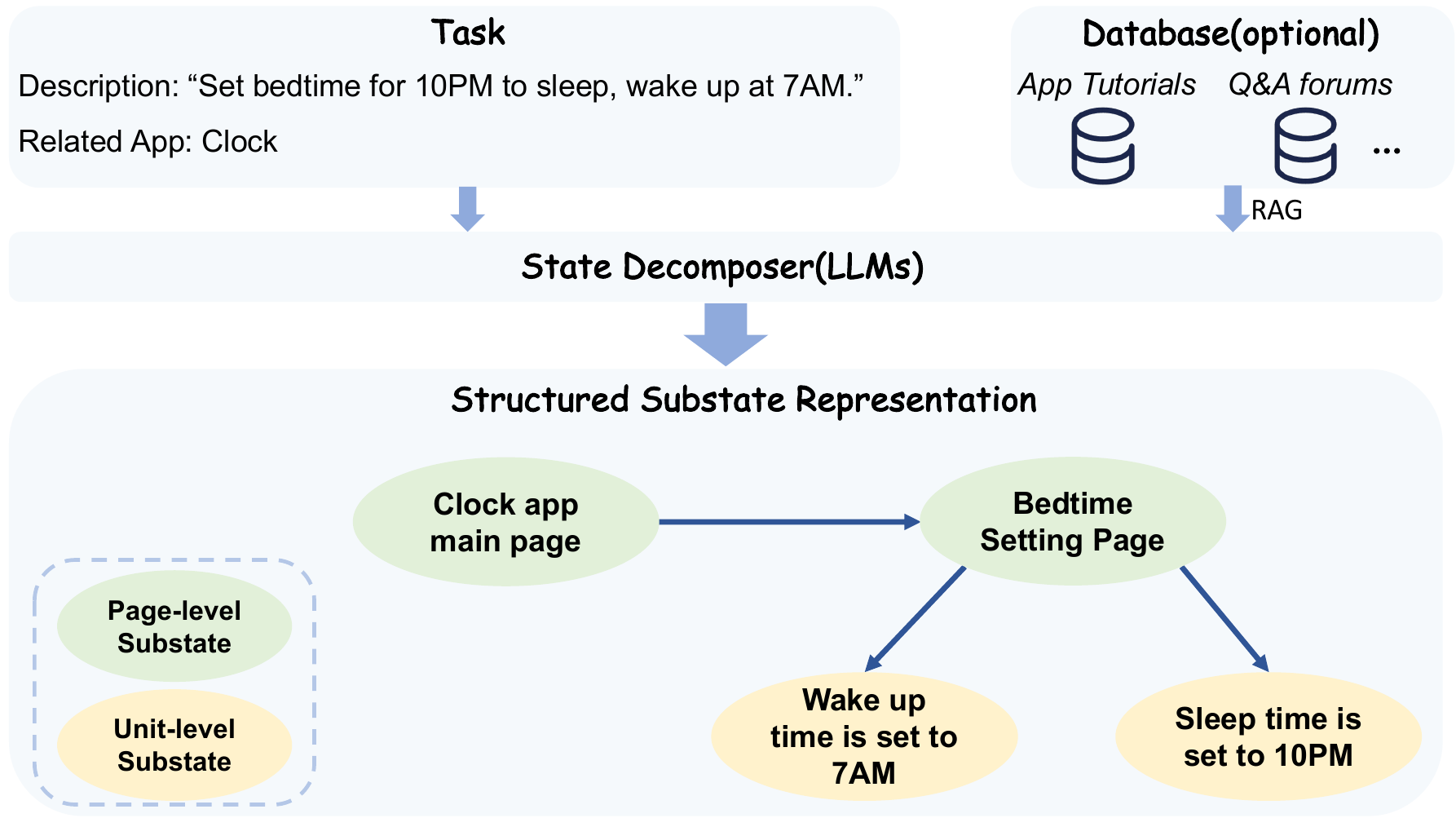}
\caption{The process of automatically generating task-specific Structured Substate Representation with State Decomposer.}
\label{fig:decomposer}
\end{figure}

Figure \ref{fig:decomposer} illustrates the process of generating substates with State Decomposer. It incorporates a pre-defined prompt that describes the SSR model and guides LLM to output substates corresponding to a specific task.

% Our key insight is leveraging LLM's knowledge of common apps gained from pre/post-training to generate substates for a specific app task. 
% For a new app that LLM has never been trained on, it can still generate accurate substates through transfer of knowledge from past experiences with similar apps.
% Additionally, we can supplement this with app-specific knowledge through Retrieval-Augmented Generation (RAG) when needed.
Our key insight is leveraging LLM's pre-trained knowledge of common apps to generate substates for specific tasks. 
Even for new apps, LLMs can transfer knowledge from similar apps to generate accurate substates, supplemented by app-specific RAG when needed.

Using the State Decomposer, we augment existing benchmarks \cite{xu2024androidlab} with autonomously generated substates to create an evaluation suite of 93 tasks.

% The State Decomposer leverages LLM's knowledge of common apps gained from pre/post-training to generate substates for a specific app task. 
% To achieve this, the State Decomposer incorporates a pre-defined prompt that describes the SSR model and guides LLM to output in correct syntax.
% % Describe why our model can work
% For a new app that LLM has never been trained on, it can still generate roughly accurate substates through transfer of knowledge from past experiences with similar apps.
% Moreover, we can optionally provide a small amount of extra knowledge of a new app to help LLM generate more accurate substates.

% Leveraging the State Decomposer, we augment existing benchmark \cite{xu2024androidlab} with autonomously generated substates, 
% resulting in a new evaluation suite comprising 93 tasks for agent evaluation.

% \subsection{Agent Observer}
% \label{subsec:Agent Observer}
% The Agent Observer is designed to run agent as black-box with a task and capture its screenshots-trajectory during task execution.
% Specifically, we first adapt existing mobile agents to implement a general interface that defines how to run the agent, each requires only a few additional lines of code.
% After that, the Agent Observer can launch these agents transparently, utilizes tools like Android Debug Bridge (adb) to capture screenshots, filters out the duplicate ones, and saves them as screenshots-trajectory for later evaluation.

\label{subsec:Judge System}
\subsection{Judge System}

\begin{figure}[htbp]
\centering
\includegraphics[width=0.8\columnwidth]{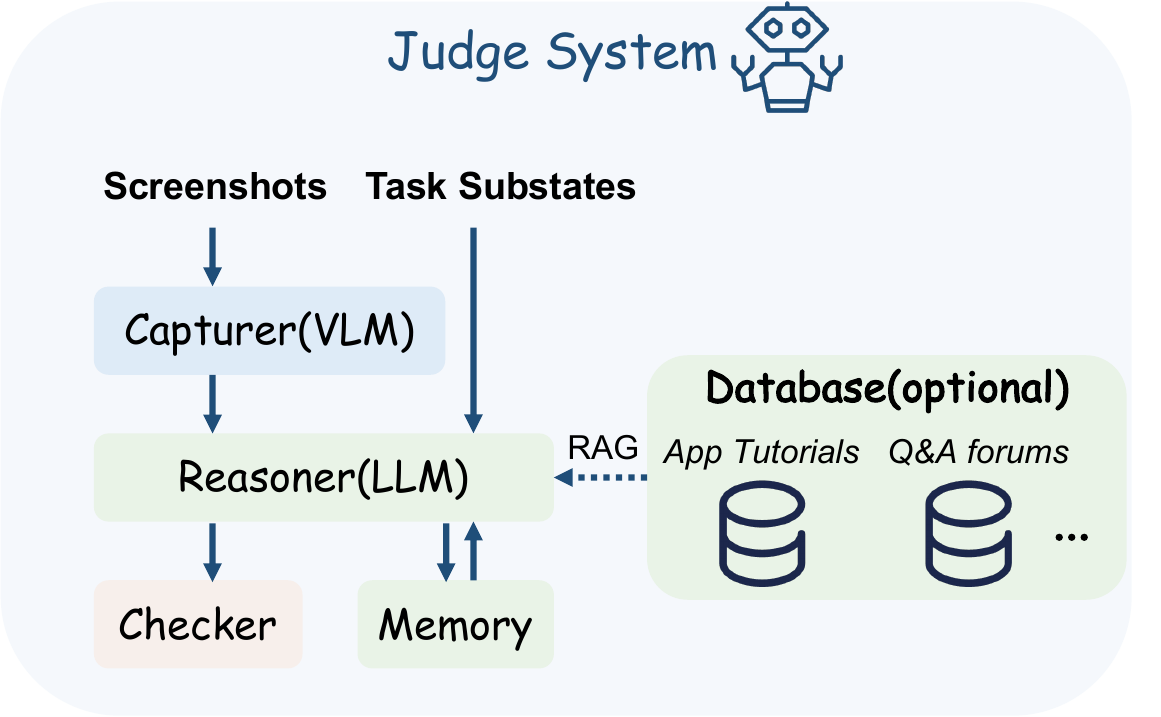}
\caption{The architecture of Judge System.}
\label{fig:judge}
\end{figure}

The architecture of Judge System is shown in Figure \ref{fig:judge}.
Given the substates of a task and the screenshots-trajectory during the agent's task execution,
Judge System autonomously evaluates the agent's functionality correctness with substate-level feedback.
It has three key components: Capturer, Reasoner and Checker.
The overall judging process is summarized in Algorithm \ref{alg:judge}. We provide a detailed description of each component in the following.
\begin{algorithm}[hbt!]
\caption{Judging Process}\label{alg:judge}
\begin{algorithmic}
\REQUIRE 
    \STATE Task description $t$, Task substates $\mathcal{S}_t$, Screenshots-trajectory $\mathcal{SC}_t$
\STATE set $mem \leftarrow []$
\FOR{each screenshot $sc_i \in \mathcal{SC}_t$}
    \STATE $d_i = \text{Capturer}(sc_i)$
    \STATE $r_i, critical\_info = \text{Reasoner}(d_i, \mathcal{S}_t, mem)$
    \STATE Append $critical\_info$ to $mem$
    \STATE $\mathcal{S}_t \leftarrow \text{Checker}(r_i)$
\ENDFOR
\end{algorithmic}
\end{algorithm}

\textbf{Capturer}. 
The Capturer is built upon a Vision Language Model (VLM) that converts screenshots into detailed textual descriptions of layout, content, and app identification.
% Instead of utilizing an end-to-end VLM doing complex image understanding and reasoning simultaneously,
% a dedicated Capturer generates only the screenshot description, which places low demands on VLM's capability while providing more controllability.

\textbf{Reasoner}. 
The Reasoner is based on a Large Language Model (LLM) and plays a crucial role in judging an agent's performance.
Given the description of the ith screenshot $d_i$, task $t$, task substates representation $\mathcal{S}_t$,
Reasoner generates an analysis $a_i$ and a judge result $j_i$ for each substate $s_i \in \mathcal{S}_t$.
The judge result $j_i$ for substate $s_i$ is either Success if $d_i$ matches $s_i$ or Uncertain otherwise.

Integrated with the structured substates representation, we can prompt the Reasoner to reason about each of substates following Algorithm \ref{alg:reasoner}.
We enhance Reasoner with a memory module storing critical information the Reasoner proactively reported and successful substates, and optionally use Retrieval-Augmented Generation (RAG) to incorporate app-specific knowledge for more accurate results.

\begin{algorithm}[hbt!]
\caption{Reasoning Process}\label{alg:reasoner}
\begin{algorithmic}
\REQUIRE Current screenshot description $d_i$, substates to check $\mathcal{S}_t$, reasoner's memory $mem$
\STATE $N \leftarrow ||\mathcal{S}_t||$
\STATE $\text{Initialize } j_i \leftarrow Uncertain \text{ for all } i \in N$
\STATE $res \leftarrow []$
\FOR{$s_i \in \mathcal{S}_t$}
    \STATE $\text{critical\_info}, a_i, j_i \leftarrow \text{Match}(d_i, s_i, mem)$
    \IF{$s_i$ is a UnitNode}
        \STATE $j_i \leftarrow j_{s_i.parent} \wedge j_i$
    \ENDIF
    \STATE Append $j_i$ to $res$
\ENDFOR
\end{algorithmic}
\end{algorithm}

\textbf{Checker}.
During our empirical evaluation, we find that the Reasoner occasionally generates unexpected outputs, failing to strictly follow the reasoning process in Algorithm~\ref{alg:reasoner}.
So, we add a Checker module to guarantee the Reasoner's output is acceptable and consistent with the reasoning process.
There are two rules that Checker guarantees: 
1) Each substate judge result must be either Success or Uncertain.
2) UnitNodes can only be judged as Success if their parent PageNode is also Success.
If violated, Checker retries or skips the current judgment.

%-------------------------------------------------------------------------------
\section{Evaluation}
We evaluate \sys{} on a variety of popular Large Language Models (LLMs) configurations and mobile agents.
Our evaluation seeks to answer the following questions:
\begin{enumerate}[itemsep=1pt,topsep=2pt,partopsep=1pt]
\setlength{\itemsep}{0pt}
\setlength{\parskip}{0pt}
\setlength{\parsep}{0pt}
    \item[\textbf{Q1.}] What is the accuracy and completeness of the substates generated by State Decomposer (Section \ref{subsec:state-decomposer}) compared to human-annotated substates?
    \item[\textbf{Q2.}] What is the accuracy of the Judge System (Section \ref{subsec:Judge System})?
    \item[\textbf{Q3.}] Can our framework effectively demonstrate the fine-grained and comprehensive performance of mobile agents?
\end{enumerate}

\textbf{Models and Environments}. 
We evaluate \sys{} across multiple LLM configurations, including GPT-4o\cite{hurst2024gpt}, DeepSeek V3\cite{liu2024deepseek}, Gemini-2.0-flash-thinking\cite{team2023gemini}. 
The Capturer in Judge System utilizes Gemini-2.0-flash as its backend VLM.
We conduct our experiments on an Android Virtual Device (AVD) emulator as the mobile environment for agent execution. 
The emulator runs Android 13 (API level 33) with a high-resolution display (1440x3120 pixels).

\textbf{Mobile Agent}. 
% The setup of the environment can be further automated by tools like \cite{android-uiautomator2}.
We evaluate a prompt-based agent and a training-based agent using our framework.
Each agent starts performing tasks in an identical manually initialized environment. 
\vspace{-0.4em}
\begin{itemize}
\setlength{\itemsep}{0pt}
\setlength{\parskip}{0pt}
\setlength{\parsep}{0pt}
    \item Mobile-Agent-E \cite{wang2025mobile}: Hierarchical multi-agent framework.
    We select Qwen-VL-Plus \cite{Bai2023QwenVLAV} as its caption model following the original paper and use gemini-2.0-flash-thinking as its reasoning model.
    \item CogAgent \cite{Hong2023CogAgentAV}: Training-based agent.
    In our evaluation, we test on the latest version of the CogAgent-9B-20241220. 
    % Although CogAgent natively supports operations on the Android platform, it doesn't officially provide a framework for interacting with Android devices.
    % So, we adapt CogAgent to operate on Android devices according to its official action space documentation.
\end{itemize}
\vspace{-0.4em}

\subsection{State Decomposer Evaluation}
\label{subsec:substate-decomposer-eval}
% --------------
\begin{table*}[htbt]
\centering
\begin{tabular*}{\textwidth}{@{\extracolsep{\fill}}lcccc@{}}
\toprule
Model & Cover Rate & Redundant Rate & Optional Rate & Incorrect Rate \\
\midrule
GPT-4o & 93.28\% & 10.49\% & 9.82\% & 1.56\% \\
DeepSeek V3 & 93.94\% & 19.85\% & 8.13\% & 1.70\% \\
\bottomrule
\end{tabular*}
\caption{Evaluating State Decomposer's performance on substate generation with different Large Language Models configurations. 
We use human-annotated substates as references and compare them with automatically generated substates.
}
\label{tab:substate-eval}
\end{table*}

To answer Q1, we collect 93 tasks from the following benchmark and augment these tasks with substates using State Decomposer. 
We then manually evaluate the quality of substates generated by State Decomposer.

\textbf{AndroidLab}\cite{xu2024androidlab}: An open-source Android agent benchmark emphasizing generalizability. We evaluate all Operation Tasks across 9 applications.

% \vspace{-0.4em}
% \begin{itemize}
% \setlength{\itemsep}{0pt}
% \setlength{\parskip}{0pt}
% \setlength{\parsep}{0pt}
%     \item \textbf{AndroidLab:} An open-source Android agent benchmark emphasizing generalizability. We evaluate all Operation Tasks across 9 applications.
% \end{itemize}
% \vspace{-0.4em}

\textbf{Metrics.}
To evaluate the quality of automatically generated task substates ($\mathcal{S}_\text{Auto}$), 
we compare them with manually defined substates ($\mathcal{S}_\text{Human}$) using Cover Rate, Redundant Rate, Optional Rate, and Incorrect Rate metrics.
\vspace{-0.2em}
\begin{itemize}
\setlength{\itemsep}{0pt}
\setlength{\parskip}{0pt}
\setlength{\parsep}{0pt}
\setlength{\topsep}{0pt}
\setlength{\partopsep}{0pt}
    \item Cover Rate: measures how many substates in $\mathcal{S}_\text{Human}$ are covered by $\mathcal{S}_\text{Auto}$.
    \item Redundant Rate: indicates the proportion of duplicate and redundant substates in $\mathcal{S}_\text{Auto}$.
    \item Optional Rate: agents can take multiple paths to achieve the same task goal, resulting in some substates being path-dependent.
    We define these path-dependent substates in $\mathcal{S}_\text{Auto}$ as optional substates.
    \item Incorrect Rate: represents the percentage of invalid substates in $\mathcal{S}_\text{Auto}$ due to hallucinated content.
\end{itemize}
\vspace{-0.2em}

% Analysis for table 
\textbf{Results.}
Table~\ref{tab:substate-eval} compares State Decomposer performance across different LLM configurations.
Both GPT-4o and DeepSeek V3 achieve a high Cover Rate with low Incorrect Rate, indicating strong capability in identifying essential task substates. 
The impact of high Redundant Rate (19.85\%) is also minimal, as they can be treated as a single substate during evaluation.
Both models exhibit similar Optional Rate, 
likely because their knowledge is constrained to substates on all seen task completion paths, limiting their ability to identify some substates that could be non-existent in other unseen but valid task completion paths.

\begin{table}[htbp]
\centering
\setlength{\tabcolsep}{3.5pt}
\begin{tabularx}{\columnwidth}{@{}lcccccc@{}}
\toprule
\multirow{2}{*}{Model} & \multicolumn{3}{c}{Human Trace} & \multicolumn{3}{c}{Agent Trace} \\
\cmidrule(lr){2-4} \cmidrule(lr){5-7}
& SR(\%) & FP(\%) & FN(\%) & SR(\%) & FP(\%) & FN(\%) \\
\midrule
\mbox{GPT-4o} & 87.76 & 1.90 & 10.34 & 87.70 & 2.05 & 10.25 \\
\mbox{DeepSeek} & 82.91 & 3.59 & 13.50 & 90.33 & 1.61 & 8.06 \\
\mbox{Gemini} & 94.31 & 0.42 & 5.27 & 94.35 & 2.02 & 3.63 \\
\bottomrule
\end{tabularx}
\caption{Evaluating Judge System's reliability across different Reasoner base model configurations when judging human and agent traces. 
DeepSeek-V3 is abbreviated as DeepSeek, and Gemini-2.0-flash-thinking is abbreviated as Gemini in the table.
}
\label{tab:judge-agent-comparison}
\end{table}
\subsection{Judge System Performance Evaluation}
To answer Q2, we collect 93 screenshots-trajectory traces from both human participants and Mobile-Agent-E during task execution.
These traces are then autonomously judged by the Judge System with task-specific substates generated in Section \ref{subsec:substate-decomposer-eval}.
We evaluate Judge System reliability through human verification, comparing its substate-level judgements against the actual screenshots-trajectory traces using three metrics: Judge Success Rate (SR), False Positive Rate (FP), and False Negative Rate (FN).

\textbf{Results.}
Table~\ref{tab:judge-agent-comparison} shows the accuracy of Judge System in judging both human and agent traces, while Reasoner is configured with different models.
Contributing to our design of the Judge System and Structured Substates Representation, 
we find that Judge System achieves high accuracy that aligns with human judgements.
The Gemini-2.0-flash-thinking based Judge System achieves the highest accuracy (94.35\%) among all LLM configurations, 
demonstrating the strong capability of reasoning models.

\textbf{Error Analysis.}
% We further analyze the error cases (FP/FN) in autonomous judgements and identify the following two main reasons:
We further analyze error cases in autonomous judgements and identify two main causes:
1) \textit{Limited Capturer model capabilities.}
Though Capturer performs well in most cases, it still faces the problem of information loss when converting screenshots into textual descriptions.
Missing important information leads to reasoning errors. 
For example, in the case of judging the substate "Sort button is visible", the Capturer may fail to identify the Sort button in its output even though the button is actually present in the screenshot, which leads to false negative judgements.
2) \textit{Limited Reasoner model capabilities.}
The Judge System can make incorrect judgements even given all necessary information due to the limited capabilities of reasoning and instruction following.
% While judging substate "Sort by duration option in ascending order is selected", the Reasoner can not deduce that the absence of a descending sort selection implies ascending order is active.

% Cost Analysis
% \begin{table}[htbp]
% \small
% \centering
% \begin{tabular}{lllll}
% \toprule
% \multirow{2}{*}{Component} & \multicolumn{2}{c}{Input} & \multicolumn{2}{c}{Output} \\
% \cmidrule(lr){2-3} \cmidrule(lr){4-5}
% & Tokens & Cost (\$) & Tokens & Cost (\$) \\
% \midrule
% Reasoner & 2,668 & 0.0011 & 750 & 0.0003 \\
% Capturer & 2,981 & 0.0012 & 466 & 0.0002 \\
% \bottomrule
% \end{tabular}
% \caption{
%     Average token consumption and cost per judging round for the Judge System. 
%     Reasoner builds upon Gemini-2.0-flash-thinking while Capturer uses Gemini-2.0-flash as its backend VLM.
% }
% \label{tab:cost-analysis}
% \end{table}

% \textbf{Cost Analysis.} 
% Table~\ref{tab:cost-analysis} presents the per-round API costs for the two LM-based components in the Judge System.
% On average, the Judge System requires 8 rounds to complete the evaluation of a single agent execution.
% Therefore, the total cost for evaluating one task execution amounts to approximately 0.0224 USD.

\subsection{Mobile Agent Performance Evaluation}
To answer Q3, we evaluate the absolute capability of different mobile agents using our framework on 49 tasks with substates generated in Section \ref{subsec:substate-decomposer-eval}.
We choose Gemini-2.0-flash-thinking as a backend model for Reasoner in Judge System.

\textbf{Metrics.} 
\sys{} provides substate-level evaluation for each task execution, enabling us to evaluate mobile agent performance using Substate Completion Rate (SCR) and Task Completion Rate (TCR).
\vspace{-0.5em}
\begin{itemize}
\setlength{\itemsep}{0pt}
\setlength{\parskip}{0pt}
\setlength{\parsep}{0pt}
    \item Substate Completion Rate (SCR): The average percentage of successfully completed substates across all tasks.
    \item Task Completion Rate (TCR): The percentage of tasks where all substates are successfully completed.
\end{itemize}
\vspace{-0.5em}

\begin{table}[htbp]
\centering
\begin{tabular}{lll}
\toprule
\textbf{Agent} & \multicolumn{1}{c}{\textbf{SCR (\%)}} & \multicolumn{1}{c}{\textbf{TCR (\%)}} \\
\midrule
CogAgent        & 62.59 & 22.45 \\
Mobile-Agent-E & 77.84 & 32.65 \\
\bottomrule
\end{tabular}
\caption{Performance comparison of different mobile agents.}
\label{tab:agent-comparison}
\end{table}

\textbf{Results.}
Table~\ref{tab:agent-comparison} shows the performance of different mobile agents. 
We find a notable gap between SCR and TCR for both agents (45.19\% for Mobile-Agent-E and 40.14\% for CogAgent), 
indicating that our evaluation framework can give a more realistic and fine-grained evaluation of agent performance.
The substate completion feedback from \sys{} reveals mobile agent weaknesses, 
including insufficient app-specific knowledge and action space limitations (e.g., inability to handle certain UI interactions or gestures).
This detailed substate feedback can further improve agent performance through targeted model fine-tuning or strategic prompt engineering.

%-------------------------------------------------------------------------------
\section{Conclusion}
In this study, we introduced \sys{}, a practical evaluation framework that enables autonomous fine-grained evaluation of mobile agents.
By proposing a Structured Substate Representation model, \sys{} can automatically generate reward signals for a given task.
And \sys{} designs a three-stage Judge System to autonomously evaluate the performance of mobile agents.
Our evaluation shows that \sys{} generates reward signals correlating with human annotations and achieves autonomous evaluation accuracy comparable to human evaluation.

%-------------------------------------------------------------------------------
% Limitation is required before reference according to https://aclrollingreview.org/cfp#instructions-for-two-way-anonymized-review
% \input{limitation.tex}

%-------------------------------------------------------------------------------
\bibliographystyle{IEEEbib}
\bibliography{custom}

\begin{thebibliography}{10}

\bibitem{Zhang2023AppAgentMA}
Chi Zhang, Zhao Yang, Jiaxuan Liu, Yanda Li, Yucheng Han, Xin Chen, Zebiao
  Huang, Bin Fu, and Gang Yu,
\newblock ``Appagent: Multimodal agents as smartphone users,''
\newblock in {\em Proceedings of the 2025 CHI Conference on Human Factors in
  Computing Systems}, 2025, pp. 1--20.

\bibitem{zhang-zhang-2024-look}
Zhuosheng Zhang and Aston Zhang,
\newblock ``You only look at screens: Multimodal chain-of-action agents,''
\newblock in {\em Findings of the Association for Computational Linguistics ACL
  2024}, 2024, pp. 3132--3149.

\bibitem{Nong2024MobileFlowAM}
Songqin Nong, Jiali Zhu, Rui Wu, Jiongchao Jin, Shuo Shan, Xiutian Huang, and
  Wenhao Xu,
\newblock ``Mobileflow: A multimodal llm for mobile gui agent,''
\newblock {\em arXiv preprint arXiv:2407.04346}, 2024.

\bibitem{Lee2023ExploreSD}
Sunjae Lee, Junyoung Choi, Jungjae Lee, Munim~Hasan Wasi, Hojun Choi, Steven~Y
  Ko, Sangeun Oh, and Insik Shin,
\newblock ``Explore, select, derive, and recall: Augmenting llm with human-like
  memory for mobile task automation,''
\newblock {\em arXiv preprint arXiv:2312.03003}, 2023.

\bibitem{Wen2024AutoDroid}
Hao Wen, Yuanchun Li, Guohong Liu, Shanhui Zhao, Tao Yu, Toby Jia-Jun Li, Shiqi
  Jiang, Yunhao Liu, Yaqin Zhang, and Yunxin Liu,
\newblock ``Autodroid: Llm-powered task automation in android,''
\newblock in {\em Proceedings of the 30th Annual International Conference on
  Mobile Computing and Networking}, 2024, pp. 543--557.

\bibitem{Wang2024MobileAgentv2MD}
Junyang Wang, Haiyang Xu, Haitao Jia, Xi~Zhang, Ming Yan, Weizhou Shen,
  Ji~Zhang, Fei Huang, and Jitao Sang,
\newblock ``Mobile-agent-v2: Mobile device operation assistant with effective
  navigation via multi-agent collaboration,''
\newblock {\em Advances in Neural Information Processing Systems}, vol. 37, pp.
  2686--2710, 2024.

\bibitem{Hong2023CogAgentAV}
Wenyi Hong, Weihan Wang, Qingsong Lv, Jiazheng Xu, Wenmeng Yu, Junhui Ji, Yan
  Wang, Zihan Wang, Yuxiao Dong, Ming Ding, et~al.,
\newblock ``Cogagent: A visual language model for gui agents,''
\newblock in {\em Proceedings of the IEEE/CVF Conference on Computer Vision and
  Pattern Recognition}, 2024, pp. 14281--14290.

\bibitem{wang2025mobile}
Zhenhailong Wang, Haiyang Xu, Junyang Wang, Xi~Zhang, Ming Yan, Ji~Zhang, Fei
  Huang, and Heng Ji,
\newblock ``Mobile-agent-e: Self-evolving mobile assistant for complex tasks,''
\newblock {\em arXiv preprint arXiv:2501.11733}, 2025.

\bibitem{qin2025ui}
Yujia Qin, Yining Ye, Junjie Fang, Haoming Wang, Shihao Liang, Shizuo Tian,
  Junda Zhang, Jiahao Li, Yunxin Li, Shijue Huang, et~al.,
\newblock ``Ui-tars: Pioneering automated gui interaction with native agents,''
\newblock {\em arXiv preprint arXiv:2501.12326}, 2025.

\bibitem{li2020mapping}
Yang Li, Jiacong He, Xin Zhou, Yuan Zhang, and Jason Baldridge,
\newblock ``Mapping natural language instructions to mobile ui action
  sequences,''
\newblock {\em arXiv preprint arXiv:2005.03776}, 2020.

\bibitem{Rawles2023AndroidIW}
Christopher Rawles, Alice Li, Daniel Rodriguez, Oriana Riva, and Timothy
  Lillicrap,
\newblock ``Androidinthewild: A large-scale dataset for android device
  control,''
\newblock {\em Advances in Neural Information Processing Systems}, vol. 36, pp.
  59708--59728, 2023.

\bibitem{xing2024androidarena}
Mingzhe Xing, Rongkai Zhang, Hui Xue, Qi~Chen, Fan Yang, and Zhen Xiao,
\newblock ``Understanding the weakness of large language model agents within a
  complex android environment,''
\newblock in {\em Proceedings of the 30th ACM SIGKDD Conference on Knowledge
  Discovery and Data Mining}, 2024, pp. 6061--6072.

\bibitem{zhang2024llamatouch}
Li~Zhang, Shihe Wang, Xianqing Jia, Zhihan Zheng, Yunhe Yan, Longxi Gao,
  Yuanchun Li, and Mengwei Xu,
\newblock ``Llamatouch: A faithful and scalable testbed for mobile ui task
  automation,''
\newblock in {\em Proceedings of the 37th Annual ACM Symposium on User
  Interface Software and Technology}, 2024, pp. 1--13.

\bibitem{li2024effects}
Wei Li, William~E Bishop, Alice Li, Christopher Rawles, Folawiyo
  Campbell-Ajala, Divya Tyamagundlu, and Oriana Riva,
\newblock ``On the effects of data scale on ui control agents,''
\newblock {\em Advances in Neural Information Processing Systems}, vol. 37, pp.
  92130--92154, 2024.

\bibitem{venkatesh2022ugif}
Sagar~Gubbi Venkatesh, Partha Talukdar, and Srini Narayanan,
\newblock ``Ugif: Ui grounded instruction following,''
\newblock {\em arXiv preprint arXiv:2211.07615}, 2022.

\bibitem{klissarov2023motif}
Martin Klissarov, Pierluca D'Oro, Shagun Sodhani, Roberta Raileanu, Pierre-Luc
  Bacon, Pascal Vincent, Amy Zhang, and Mikael Henaff,
\newblock ``Motif: Intrinsic motivation from artificial intelligence
  feedback,''
\newblock {\em arXiv preprint arXiv:2310.00166}, 2023.

\bibitem{xu2024androidlab}
Yifan Xu, Xiao Liu, Xueqiao Sun, Siyi Cheng, Hao Yu, Hanyu Lai, Shudan Zhang,
  Dan Zhang, Jie Tang, and Yuxiao Dong,
\newblock ``Androidlab: Training and systematic benchmarking of android
  autonomous agents,''
\newblock {\em arXiv preprint arXiv:2410.24024}, 2024.

\bibitem{rawles2024androidworld}
Christopher Rawles, Sarah Clinckemaillie, Yifan Chang, Jonathan Waltz,
  Gabrielle Lau, Marybeth Fair, Alice Li, William Bishop, Wei Li, Folawiyo
  Campbell-Ajala, et~al.,
\newblock ``Androidworld: A dynamic benchmarking environment for autonomous
  agents,''
\newblock {\em arXiv preprint arXiv:2405.14573}, 2024.

\bibitem{Pan2024AutonomousEA}
Jiayi Pan, Yichi Zhang, Nicholas Tomlin, Yifei Zhou, Sergey Levine, and Alane
  Suhr,
\newblock ``Autonomous evaluation and refinement of digital agents,''
\newblock {\em arXiv preprint arXiv:2404.06474}, 2024.

\bibitem{Branavan2009ReinforcementLF}
Satchuthananthavale~RK Branavan, Harr Chen, Luke Zettlemoyer, and Regina
  Barzilay,
\newblock ``Reinforcement learning for mapping instructions to actions,''
\newblock in {\em Proceedings of the Joint Conference of the 47th Annual
  Meeting of the ACL and the 4th International Joint Conference on Natural
  Language Processing of the AFNLP}, 2009, pp. 82--90.

\bibitem{Shvo2021AppBuddyLT}
Maayan Shvo, Zhiming Hu, Rodrigo~Toro Icarte, Iqbal Mohomed, Allan Jepson, and
  Sheila~A McIlraith,
\newblock ``Appbuddy: Learning to accomplish tasks in mobile apps via
  reinforcement learning,''
\newblock {\em arXiv preprint arXiv:2106.00133}, 2021.

\bibitem{Gur2022EnvironmentGF}
Izzeddin Gur, Natasha Jaques, Yingjie Miao, Jongwook Choi, Manoj Tiwari,
  Honglak Lee, and Aleksandra Faust,
\newblock ``Environment generation for zero-shot compositional reinforcement
  learning,''
\newblock {\em Advances in Neural Information Processing Systems}, vol. 34, pp.
  4157--4169, 2021.

\bibitem{li-etal-2020-mapping}
Yang Li, Jiacong He, Xin Zhou, Yuan Zhang, and Jason Baldridge,
\newblock ``Mapping natural language instructions to mobile ui action
  sequences,''
\newblock {\em arXiv preprint arXiv:2005.03776}, 2020.

\bibitem{Humphreys2022ADA}
Peter~C Humphreys, David Raposo, Tobias Pohlen, Gregory Thornton, Rachita
  Chhaparia, Alistair Muldal, Josh Abramson, Petko Georgiev, Adam Santoro, and
  Timothy Lillicrap,
\newblock ``A data-driven approach for learning to control computers,''
\newblock in {\em International Conference on Machine Learning}. PMLR, 2022,
  pp. 9466--9482.

\bibitem{Taeb2023AXNavRA}
Maryam Taeb, Amanda Swearngin, Eldon Schoop, Ruijia Cheng, Yue Jiang, and
  Jeffrey Nichols,
\newblock ``Axnav: Replaying accessibility tests from natural language,''
\newblock in {\em Proceedings of the 2024 CHI Conference on Human Factors in
  Computing Systems}, 2024, pp. 1--16.

\bibitem{Lu2024OmniParserFP}
Yadong Lu, Jianwei Yang, Yelong Shen, and Ahmed Awadallah,
\newblock ``Omniparser for pure vision based gui agent,''
\newblock {\em arXiv preprint arXiv:2408.00203}, 2024.

\bibitem{Lu2024GUIOA}
Quanfeng Lu, Wenqi Shao, Zitao Liu, Fanqing Meng, Boxuan Li, Botong Chen,
  Siyuan Huang, Kaipeng Zhang, Yu~Qiao, and Ping Luo,
\newblock ``Gui odyssey: A comprehensive dataset for cross-app gui navigation
  on mobile devices,''
\newblock {\em arXiv preprint arXiv:2406.08451}, 2024.

\bibitem{Bai2024DigiRLTI}
Hao Bai, Yifei Zhou, Jiayi Pan, Mert Cemri, Alane Suhr, Sergey Levine, and
  Aviral Kumar,
\newblock ``Digirl: Training in-the-wild device-control agents with autonomous
  reinforcement learning,''
\newblock {\em Advances in Neural Information Processing Systems}, vol. 37, pp.
  12461--12495, 2024.

\bibitem{Li2023AZL}
Tao Li, Gang Li, Zhiwei Deng, Bryan Wang, and Yang Li,
\newblock ``A zero-shot language agent for computer control with structured
  reflection,''
\newblock {\em arXiv preprint arXiv:2310.08740}, 2023.

\bibitem{Wu2024OSCopilotTG}
Zhiyong Wu, Chengcheng Han, Zichen Ding, Zhenmin Weng, Zhoumianze Liu, Shunyu
  Yao, Tao Yu, and Lingpeng Kong,
\newblock ``Os-copilot: Towards generalist computer agents with
  self-improvement,''
\newblock {\em arXiv preprint arXiv:2402.07456}, 2024.

\bibitem{Shinn2023ReflexionLA}
Noah Shinn, Federico Cassano, Ashwin Gopinath, Karthik Narasimhan, and Shunyu
  Yao,
\newblock ``Reflexion: Language agents with verbal reinforcement learning,''
\newblock {\em Advances in Neural Information Processing Systems}, vol. 36, pp.
  8634--8652, 2023.

\bibitem{Zhuge2024AgentasaJudgeEA}
Mingchen Zhuge, Changsheng Zhao, Dylan Ashley, Wenyi Wang, Dmitrii Khizbullin,
  Yunyang Xiong, Zechun Liu, Ernie Chang, Raghuraman Krishnamoorthi, Yuandong
  Tian, et~al.,
\newblock ``Agent-as-a-judge: Evaluate agents with agents,''
\newblock {\em arXiv preprint arXiv:2410.10934}, 2024.

\bibitem{fu-etal-2024-gptscore}
Jinlan Fu, See-Kiong Ng, Zhengbao Jiang, and Pengfei Liu,
\newblock ``Gptscore: Evaluate as you desire,''
\newblock {\em arXiv preprint arXiv:2302.04166}, 2023.

\bibitem{Chen2024MLLMasaJudgeAM}
Dongping Chen, Ruoxi Chen, Shilin Zhang, Yaochen Wang, Yinuo Liu, Huichi Zhou,
  Qihui Zhang, Yao Wan, Pan Zhou, and Lichao Sun,
\newblock ``Mllm-as-a-judge: Assessing multimodal llm-as-a-judge with
  vision-language benchmark,''
\newblock in {\em Forty-first International Conference on Machine Learning},
  2024.

\bibitem{Zhao2024AutoArenaAL}
Ruochen Zhao, Wenxuan Zhang, Yew~Ken Chia, Weiwen Xu, Deli Zhao, and Lidong
  Bing,
\newblock ``Auto-arena: Automating llm evaluations with agent peer battles and
  committee discussions,''
\newblock {\em arXiv preprint arXiv:2405.20267}, 2024.

\bibitem{Chan2023ChatEvalTB}
Chi-Min Chan, Weize Chen, Yusheng Su, Jianxuan Yu, Wei Xue, Shanghang Zhang,
  Jie Fu, and Zhiyuan Liu,
\newblock ``Chateval: Towards better llm-based evaluators through multi-agent
  debate,''
\newblock {\em arXiv preprint arXiv:2308.07201}, 2023.

\bibitem{Zheng2023JudgingLLM}
Lianmin Zheng, Wei-Lin Chiang, Ying Sheng, Siyuan Zhuang, Zhanghao Wu, Yonghao
  Zhuang, Zi~Lin, Zhuohan Li, Dacheng Li, Eric Xing, et~al.,
\newblock ``Judging llm-as-a-judge with mt-bench and chatbot arena,''
\newblock {\em Advances in Neural Information Processing Systems}, vol. 36, pp.
  46595--46623, 2023.

\bibitem{hurst2024gpt}
Aaron Hurst, Adam Lerer, Adam~P Goucher, Adam Perelman, Aditya Ramesh, Aidan
  Clark, AJ~Ostrow, Akila Welihinda, Alan Hayes, Alec Radford, et~al.,
\newblock ``Gpt-4o system card,''
\newblock {\em arXiv preprint arXiv:2410.21276}, 2024.

\bibitem{liu2024deepseek}
Aixin Liu, Bei Feng, Bing Xue, Bingxuan Wang, Bochao Wu, Chengda Lu, Chenggang
  Zhao, Chengqi Deng, Chenyu Zhang, Chong Ruan, et~al.,
\newblock ``Deepseek-v3 technical report,''
\newblock {\em arXiv preprint arXiv:2412.19437}, 2024.

\bibitem{team2023gemini}
Gemini Team, Rohan Anil, Sebastian Borgeaud, Jean-Baptiste Alayrac, Jiahui Yu,
  Radu Soricut, Johan Schalkwyk, Andrew~M Dai, Anja Hauth, Katie Millican,
  et~al.,
\newblock ``Gemini: a family of highly capable multimodal models,''
\newblock {\em arXiv preprint arXiv:2312.11805}, 2023.

\bibitem{Bai2023QwenVLAV}
Jinze Bai, Shuai Bai, Yunfei Chu, Zeyu Cui, Kai Dang, Xiaodong Deng, Yang Fan,
  Wenbin Ge, Yu~Han, Fei Huang, et~al.,
\newblock ``Qwen technical report,''
\newblock {\em arXiv preprint arXiv:2309.16609}, 2023.

\end{thebibliography}
% \bibliography{ref}

%%%%%%%%%%%%%%%%%%%%%%%%%%%%%%%%%%%%%%%%%%%%%%%%%%%%%%%%%%%%%%%%%%%%%%%%%%%%%%%
%%%%%%%%%%%%%%%%%%%%%%%%%%%%%%%%%%%%%%%%%%%%%%%%%%%%%%%%%%%%%%%%%%%%%%%%%%%%%%%
% APPENDIX
%%%%%%%%%%%%%%%%%%%%%%%%%%%%%%%%%%%%%%%%%%%%%%%%%%%%%%%%%%%%%%%%%%%%%%%%%%%%%%%
%%%%%%%%%%%%%%%%%%%%%%%%%%%%%%%%%%%%%%%%%%%%%%%%%%%%%%%%%%%%%%%%%%%%%%%%%%%%%%%
\newpage
\appendix
\onecolumn

\section{Prompts used in \sys{}}
In this section, we provide the prompts used in \sys{}, including the State Decomposer , Reasoner, and the Capturer Prompt.
\label{app:prompts}
\subsection{State Decomposer Prompt}
\label{app:decomposer-prompt}
% Prompts for State Decomposer
\begin{tcolorbox}[colback=red!5!white,colframe=red!75!black, enhanced, breakable]
\texttt{You are an expert agent for decomposing Android phone tasks into specific, observable states.}

\texttt{Your objective is to break down a given task into a series of clear substates that can be easily verified by examining visible on-screen information.}

\texttt{The list of substates you planned represents the ui state transition process while executing the task.}

\texttt{Example Task: Search and subscribe to the "MrBeast" YouTube channel using the YouTube app.}

\texttt{Example Task's related app: YouTube app}

\texttt{After decomposing the task, you response like the following example:}

\texttt{0. PageNode(content="Youtube main page is visible", parent\_id=None)}

\texttt{1. PageNode(content="Youtube search page is visible", parent\_id=0)}

\texttt{2. UnitNode(content="The search bar in youtube search page contains the text "MrBeast"", parent\_id=1)}

\texttt{3. PageNode(content="MrBeast channel page is visible", parent\_id=1)}

\texttt{4. UnitNode(content="MrBeast channel is subscribed", parent\_id=3)}

\texttt{REMEMBER:}

\texttt{- Represent substate by node. Node can be PageNode or UnitNode.}

\texttt{    - PageNode is a node that represents a page in the app,}

\texttt{    - UnitNode is a node that represents a unit element in its parent page like button, text, search bar etc.}

\texttt{- Each node MUST have a unique ID, which is strictly increasing.}

\texttt{    - Each node contains a content field and a parent\_id field.}

\texttt{        - In content field of each node, you should describe the ui state that should be checked as detailed as possible, content field can not be None.}

\texttt{        - In parent\_id field of each node, you should describe node parent node's id.}

\texttt{            - UnitNode's parent means the unit is in which page.}

\texttt{            - Each node's parent must be previous PageNode!}

\texttt{            - A PageNode's parent PageNode represents the page that the current page is entered from.}

\texttt{            - A UnitNode's parent PageNode represents the page in which the unit is located.}

\texttt{- Return only the list of substates without any additional information or commentary.}

\texttt{- Aim to identify the minimal number of substates needed for verification based solely on what is visible on the screen.}

\texttt{- ONLY return the key substates that are unescapable while executing the task.}

\texttt{- Remember to try your best to describe the substates as accurate as possible.}

\texttt{- Don't include any unnecessary, redundant, unclear, or similar substates!} 

\texttt{- Each substate should be as simple as possible and include only one property to be checked.}

\texttt{- You should ALWAYS check whether the target app is opened before checking any other substates.}

\texttt{- Don't check whether the button is clicked or not, this can not be judged by visual information.}

\texttt{Below are some additional information about the user task's related app:}

\texttt{\{additional\_info\}}

\texttt{Now, let's begin:}
\end{tcolorbox}

\subsection{Reasoner Prompt}
\label{app:reasoner-prompt}
% Prompts for State Reasoner
\begin{tcolorbox}[colback=red!5!white,colframe=red!75!black, enhanced, breakable]
\texttt{You are a highly specialized Android UI state verification expert. Your role is to precisely analyze and validate UI states on Android devices with exceptional attention to detail.}
\vspace{\baselineskip}

\texttt{You will receive the following inputs to perform your analysis:}

\texttt{1. Task Description: The overall user task that needs to be accomplished}

\texttt{2. Historical Context: Critical information gathered from previously analyzed screenshots}

\texttt{3. Current UI State: A detailed textual description of the current screenshot}

\texttt{4. Verification Targets: A structured list of substates that require validation}

\vspace{\baselineskip}
\texttt{Your objective is to judge whether the screenshot can match each of the substates.}

\vspace{\baselineskip}
\texttt{You'll be given an example task description like: Subscribe to the "MrBeast" YouTube channel using the YouTube app.}

\vspace{\baselineskip}
\texttt{This example task has the following substates: }

\texttt{- PageNode(state\_id=0, content="Youtube main page is visible", parent\_id=None)}

\texttt{- PageNode(state\_id=1, content="Youtube search page is visible", parent\_id=0)}

\texttt{- UnitNode(state\_id=2, content="The search bar in youtube search page contains the text "MrBeast"", parent\_id=1)}

\texttt{- PageNode(state\_id=3, content="Search results page for 'MrBeast' is visible", parent\_id=1)}

\texttt{- UnitNode(state\_id=4, content="MrBeast channel is subscribed", parent\_id=3)}

\vspace{\baselineskip}
\texttt{You will also be given a ui\_description for the screenshot, like the following:}

\vspace{\baselineskip}
\texttt{\% UI description for the screenshot}

\vspace{\baselineskip}
\texttt{You should respond like the following example:}
\texttt{\{}

\texttt{    "thought": "Screenshot shows the home page of the Youtube app, so I have to only check PageNode that describes the home page and those UnitNodes whose parent\_id is the corresponding PageNode. In current round, I can only check substate 0. For other substates, I should judge them as uncertain.",}

\texttt{    "analysis": [}

\texttt{        "For substate 0, it's a PageNode, I have to check if the Youtube main page is visible. The screenshot clearly shows the Youtube main page, so it matches the substate 0.", }

\texttt{        "For substate 1, it's a PageNode, I have to check if the Youtube search page is visible, however current screenshot shows the Youtube main page, not in the search page, so I should judge it as uncertain.", }

\texttt{        "For substate 2, it's a UnitNode, I have to first check if current page is consistent with the substate 2's parent PageNode 1, then check if the search bar contains the text 'MrBeast'. However current screenshot shows the Youtube main page, not in the search page, so I should judge it as uncertain.", }

\texttt{        "For substate 3, it's a PageNode, I have to check if the MrBeast channel page is visible. However current screenshot shows the Youtube main page, not in the MrBeast channel page, so I should judge it as uncertain.", }

\texttt{        "For substate 4, it's a UnitNode, I have to first check if current page is consistent with the substate 4's parent PageNode 3, then check if the MrBeast channel is subscribed. However current screenshot shows the Youtube main page, not in PageNode3's MrBeast channel page, so I should judge it as uncertain."}

\texttt{    ],}

\texttt{    "states": ["true", "uncertain", "uncertain", "uncertain", "uncertain"]}

\texttt{\}}

\vspace{\baselineskip}
\texttt{Remember:} 

\texttt{1. You will be called multiple times with different screenshots. So if you cannot determine a substate's status from the current knowledge, mark it as "uncertain" and wait for later screenshots in subsequent calls.}

\texttt{2. You MUST output well-formatted JSON format directly, which should be a valid JSON string, do not output any other information.}

\texttt{3. substates are divided into two types: PageNode and UnitNode. Each node represents a state. Field state\_id uniquely identifies a node. Field parent\_id is the state\_id of the parent node. UnitNode's parent MUST be a PageNode, which means the UnitNode is in the PageNode. Field content is the description of the state that we need to check.}

\texttt{    - PageNode is a node that represents a page in the app. }

\texttt{        - Process of checking PageNode: Check if the current page is the target page. If it is not the target page like in device home screen or other pages, then output this substate as "uncertain", otherwise output this substate as "true".}

\texttt{    - UnitNode is a node that represents a unit element in its parent page like button, text, search bar etc. }

\texttt{        - Process of checking UnitNode: First check if the current page is the unit's parent page, if not then output this substate as "uncertain", if yes then check if the target unit element is visible and in the correct state, if target unit is visible and in the correct state output this substate as "true", otherwise output this substate as "uncertain".}

\texttt{4. In your response's analysis list, you should provide a detailed explanation for whether each substate is matched with the screenshot. Remember to analyze each substates! Match each substate with its corresponding analysis, where the first analysis corresponds to substate 1.The number of analyses should match the number of substates}

\texttt{5. In your response's states list, you should only return string "true" or "uncertain". Each judge\_state should correspond to the analysis result of the corresponding substate.}

\texttt{    5.1 Return "true" if it is the substate is exactly matched with the screenshot visual information according to the checking process.}

\texttt{    5.2 Return "uncertain" if this substate can not judged according to the screenshot, like target unit is not visible or target page is not visible.}

\texttt{6. You can optionally contain a critical\_info field in your response, which can help you judge the "uncertain" substates in the next few screenshots, like the previous search result should be checked in the next screenshots or some video played later is the target video.}

\texttt{7. You can use information from previous judgement results as prior knowledge when evaluating the current screenshot, as they represent events that have already occurred.}

\texttt{8. You can reasonably make some deduction by considering previous substate that checked as SUCCESS. Like if you have entered a specific app, the next fews screenshots should be about this app until you enter another app.}

\vspace{\baselineskip}
\texttt{Now, let's begin:}
\end{tcolorbox}

\subsection{Capturer Prompt}
\label{app:capturer-prompt}
% Prompts for State Capturer
\begin{tcolorbox}[colback=red!5!white,colframe=red!75!black, enhanced, breakable]
\texttt{You are a specialized Android UI analyzer with expertise in converting UI screenshots into detailed textual descriptions. Your task is to provide a comprehensive and precise analysis of Android interface elements.}

\texttt{Guidelines for Analysis:}

\texttt{1. STRUCTURE AND LAYOUT}

\texttt{- First, identify the app that the screenshot belongs to.}

\texttt{- Then begin with an overview of the current screen/page}

\texttt{- Describe the hierarchical layout structure (top-to-bottom, left-to-right)}

\texttt{- Identify the main content area and any navigation elements}

\texttt{2. UI ELEMENT DETAILS}

\texttt{For each visible UI component, describe:}

\texttt{- Element type (button, text field, checkbox, etc.)}

\texttt{- Exact text content (if any)}

\texttt{- Visual properties:}

\texttt{  * Colors (background and text)}

\texttt{  * Size and positioning}

\texttt{  * Borders and shapes}

\texttt{  * Icons or images}

\texttt{- Interactive states:}

\texttt{  * Selected/unselected}

\texttt{  * Enabled/disabled}

\texttt{  * Focused/unfocused}

\texttt{  * Expanded/collapsed}

\texttt{- Accessibility properties (if visible)}

\texttt{3. CONTEXTUAL INFORMATION}

\texttt{- Identify the screen's purpose and functionality}

\texttt{- Note any system UI elements (status bar, navigation bar)}

\texttt{- Describe any visible animations or transitions}

\texttt{- Document error states or notifications}

\texttt{STRICT REQUIREMENTS:}

\texttt{- Do not make assumptions about the app identity unless explicitly shown}

\texttt{- Use precise, factual descriptions without qualifiers like "possibly" or "maybe"}

\texttt{- Document every visible UI element's state and properties}

\texttt{- Maintain a systematic top-to-bottom analysis approach}

\texttt{- Use technical terminology for UI components}

\texttt{- Include exact text strings as they appear}

\texttt{Please analyze the provided screenshot following these guidelines.}
\end{tcolorbox}

\section{Benchmark Details}
\label{app:benchmark-details}
We select all Operation tasks from the AndroidLab~\cite{xu2024androidlab} and decompose these tasks into substates with the State Decomposer.
Here we present the distribution of tasks across applications in Table~\ref{tab:task-distribution}.
Then we randomly select 49 tasks from each application and launch agents to autonomously execute them. The tasks we selected are listed in List~\ref{lst:androidlab-tasks}.

\vspace{1.3em}

\begin{table}[htb]
    \centering
    \begin{tabular}{lc}
        \toprule
        Application & Number of Tasks \\
        \midrule
        Bluecoins & 10 \\
        Calendar & 14 \\
        Cantook & 7 \\
        Clock & 21 \\
        Contacts & 11 \\
        Maps & 5 \\
        Pi Music Player & 6 \\
        Setting & 14 \\
        Zoom & 5 \\
        \midrule
        Total & 93 \\
        \bottomrule
    \end{tabular}
    \caption{Distribution of Tasks Across Applications}
    \label{tab:task-distribution}
\end{table}

\lstset{
    basicstyle=\ttfamily\small,          % Slightly larger font size
    columns=fullflexible,
    frame=single,
    breaklines=true,
    postbreak=\mbox{\textcolor{gray}{$\hookrightarrow$}\space},  % Less aggressive color
    numbers=left,                        % Add line numbers
    numberstyle=\tiny\color{gray},       % Style for line numbers
    backgroundcolor=\color{gray!10},     % Light gray background
    commentstyle=\color{green!60!black}, % Style for comments
    keywordstyle=\color{blue},          % Style for keywords
    showstringspaces=false,             % Don't show spaces in strings
    captionpos=b                        % Caption at bottom
}

\begin{lstlisting}[
    caption={Tasks randomly sampled from AndroidLab},
    label={lst:androidlab-tasks}
]
# Bluecoins
Log an expenditure of 512 CNY in the books.
For March 8, 2024, jot down an income of 3.14 CNY with 'Weixin red packet' as the note.
Adjust the expenditure on January 15, 2025, to 500 CNY.
Switch the January 1, 2025, transaction from 'expense' to 'income' and add 'Gift' as the note.
Change the type of the transaction on January 2, 2025, from 'income' to 'expense', adjust the amount to 520 CNY, and change the note to 'Wrong Operation'.

# Calendar
Arrange an event titled "homework" for me at May 21st, and set the notification time to be 10 minutes before.
Edit the event titled "work" and add a Note "computer" to it
For the event titled "work", please help me set recurrence to be daily
arrange an event titled "this day"
edit the event titled "this day", and make it repeat weekly
Help me add a note "Hello" to the event titled "Today".
Edit the event titled "exam" and make it an all-day event.

# Cantook
Delete Don Quixote from my books.
Mark Hamlet as read.
Mark the second book I recently read as unread.
Open Romeo and Juliet.
Open the category named 'Tragedies'.

# Clock
Help me set an alarm at 10:30AM tomorrow
Turn off all alarms
Delete all alarms after 2PM
Turn off the alarm at 4PM
Delete Barcelona time from clock
Set a countdown timer for 1 hour 15 minutes but do not start it
Turn on the Wake-up alarm in Bedtime
Change home time zone to Tokyo in clock
Modify silence after to 5 minutes
Close my 7:30AM alarm
Set an alarm at 3PM

# Contacts
Add John as a contacts and set his mobile phone number to be 12345678
Add a contacts whose first name is "John", last name is "Smith", mobile phone number is 12345678, and working email as 123456@gmail.com
Add a contacts whose name is Xu, set the working phone number to be 12345678 and mobile phone number to be 87654321
Add a contacts named Chen, whose company is Tsinghua University
Create a new label as work, and add Chen, Xu into it
Add a work phone number 00112233 to contacts Chen
Add birthday to Chen as 1996/10/24
Set contacts Xu's website to be abc.github.com
Edit a message to Xu, whose content is "Nice to meet you", but do not send it
Call Chen
Delete contacts Chen

# Maps
Add the address of openai to my Work place
Navigate from my location to Stanford University
Navigate from my location to University South
Navigate from my location to OpenAI
Navigate from my location to University of California, Berkeley

# Pi Music Player
Pause the currently playing song and seek to 1 minute and 27 seconds.
Play Lightship by Sonny Boy.
Sort the songs by duration time in ascending order.

# Setting
I do not want turn on wifi automatically, turn it off
Turn off my bluetooth
Turn my phone to Dark theme
Turn off Ring vibration
Check my default browser and change it to firefox
Open settings
Turn on airplane mode

# Zoom
Join meeting 0987654321, and set my name as 'Alice'. (You should not click join button, and leave it to user)
I need to join meeting 1234567890 without audio and video. (You should not click join button, and leave it to user)
Set auto connect to audio when wifi is connected in zoom settings.
\end{lstlisting}

\section{State Decomposition Results}
\label{app:state-decomposition-results}
Here we present some substates decomposition results using GPT-4o as base model for State Decomposer and Error cases.

\begin{figure*}[htb]
\centering
\includegraphics[width=\textwidth]{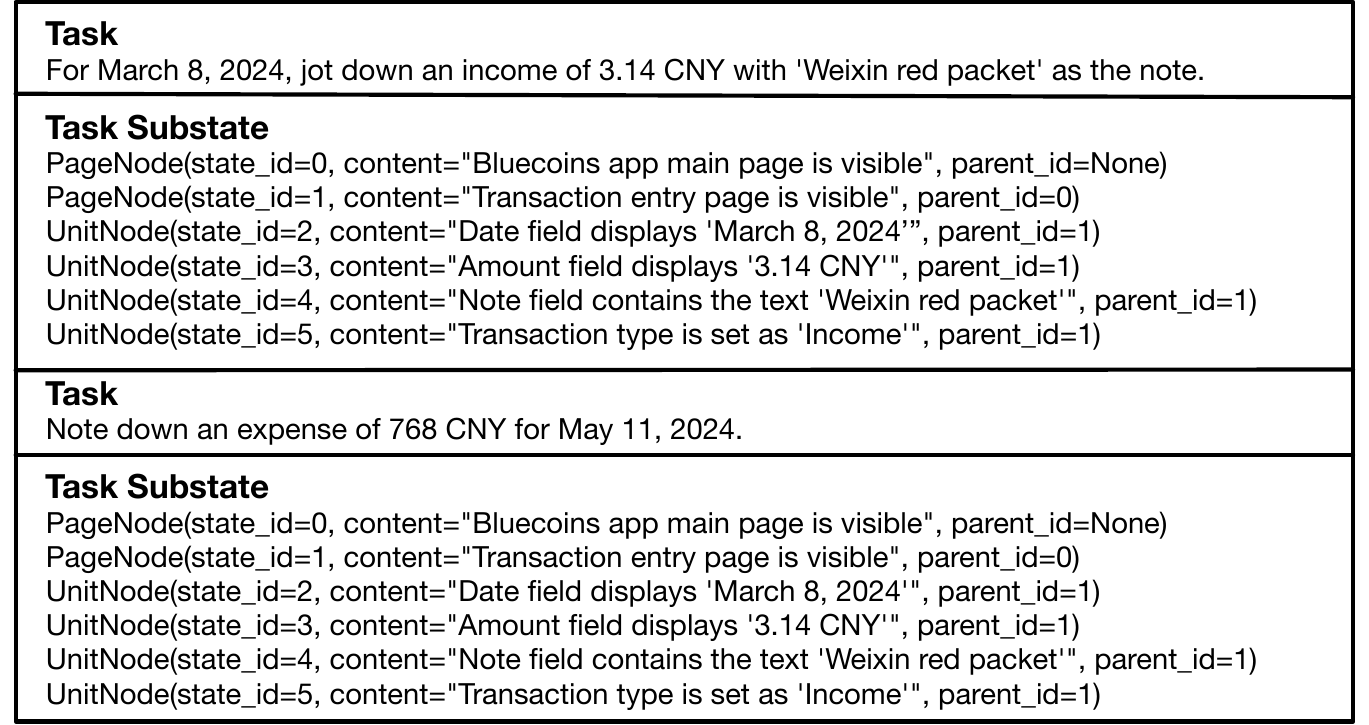}
\caption{Bluecoins task with its substates}
\vspace{-0.4cm}  % 减少与下一个图片的间距
\label{fig:bluecoins}
\end{figure*}

\begin{figure*}[htb]
\centering
\includegraphics[width=\textwidth]{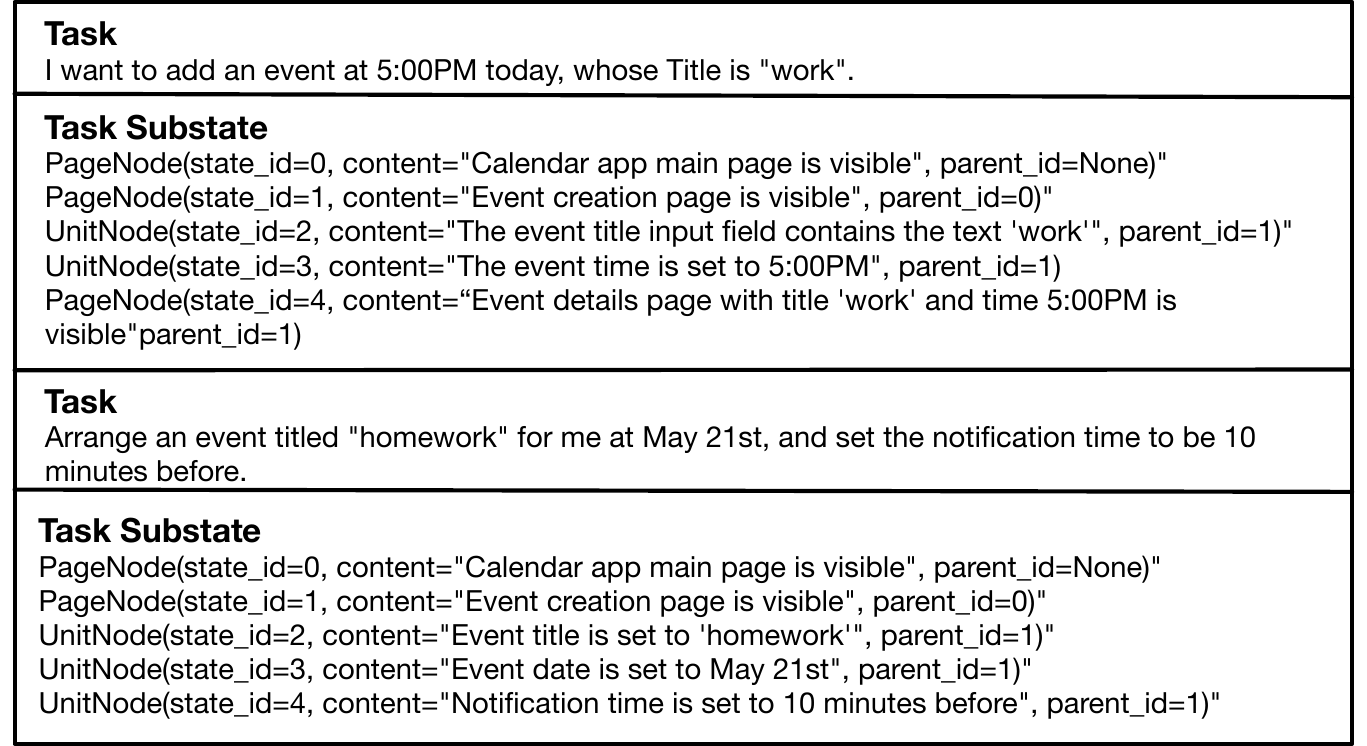}
\caption{Calendar task with its substates}
\vspace{-0.4cm}  % 减少与下一个图片的间距
\label{fig:calendar}
\end{figure*}

\begin{figure*}[htb]
\centering
\includegraphics[width=\textwidth]{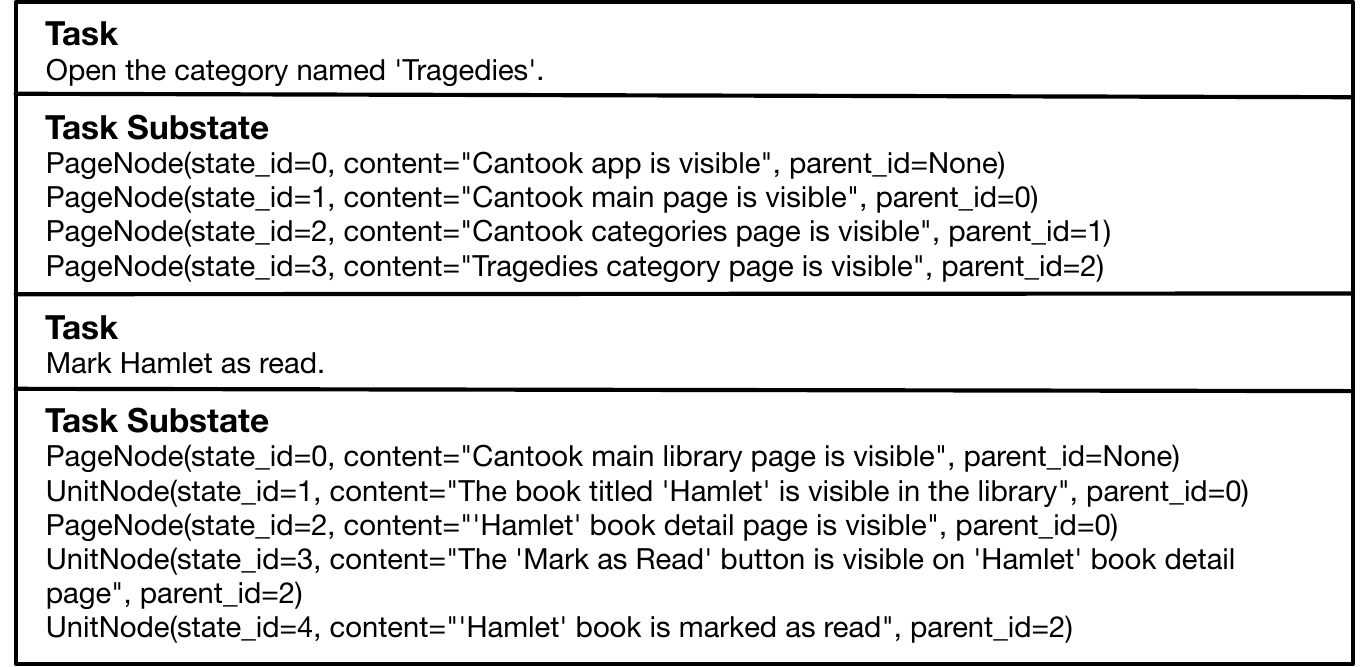}
\caption{Cantook task with its substates}
\vspace{-0.5cm}  % 减少与下一个图片的间距
\label{fig:cantook}
\end{figure*}

\begin{figure*}[htb]
\centering
\includegraphics[width=\textwidth]{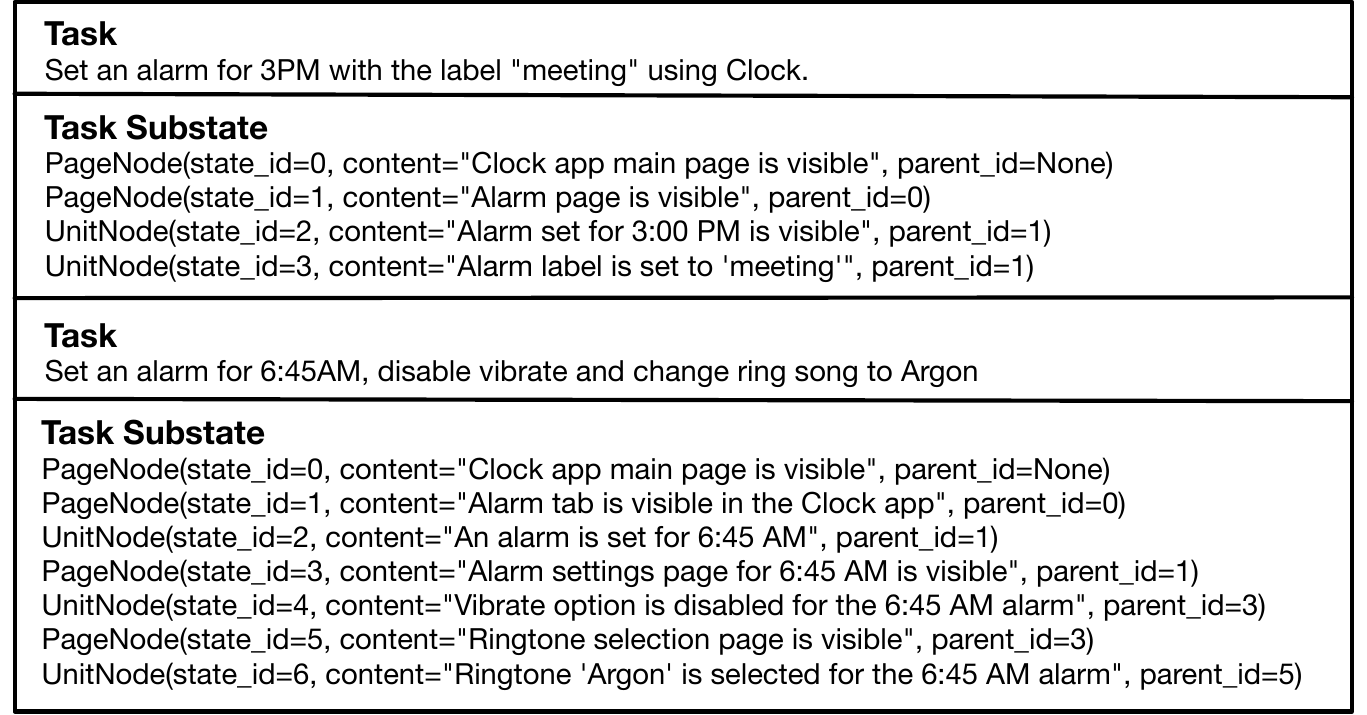}
\caption{Clock task with its substates}
\vspace{-0.4cm}  % 减少与下一个图片的间距
\label{fig:clock}
\end{figure*}

\begin{figure*}[htb]
\centering
\includegraphics[width=\textwidth]{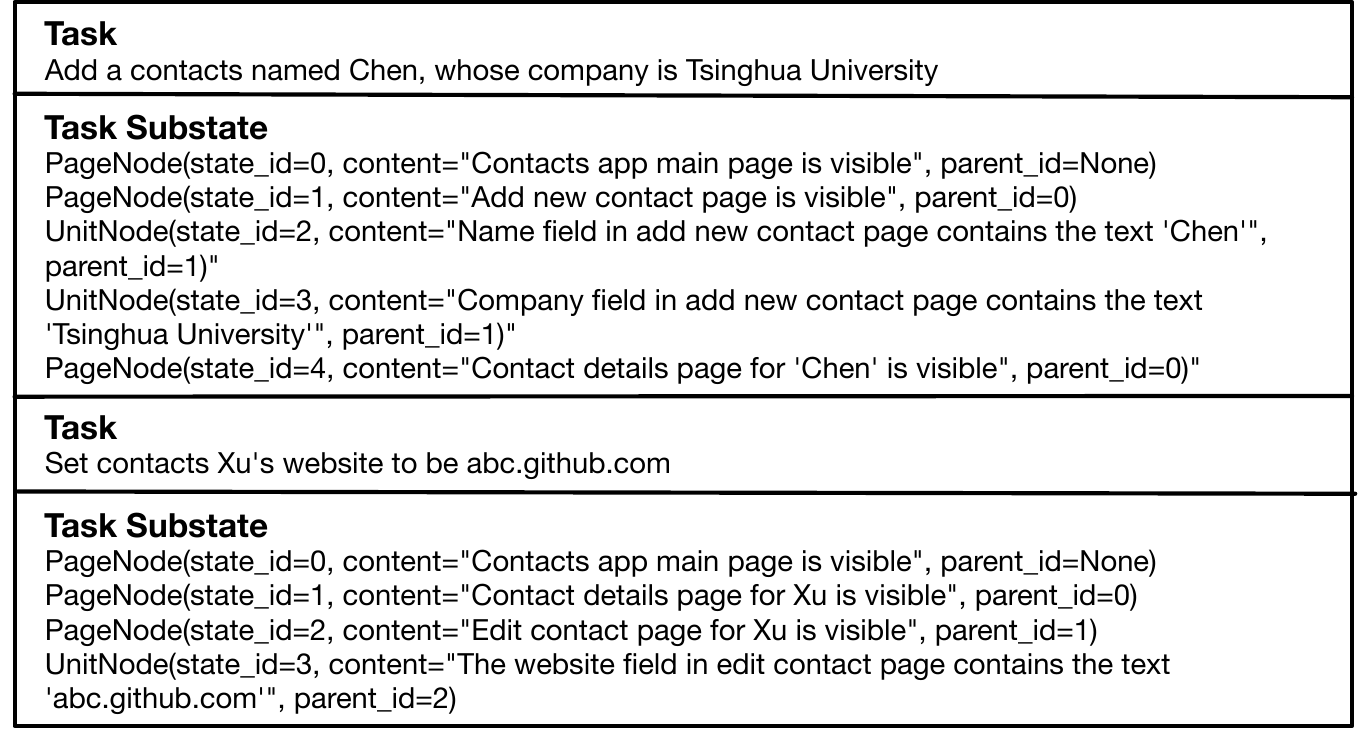}
\caption{Contacts task with its substates}
\vspace{-0.4cm}  % 减少与下一个图片的间距
\label{fig:contacts}
\end{figure*}

\begin{figure*}[htb]
\centering
\includegraphics[width=\textwidth]{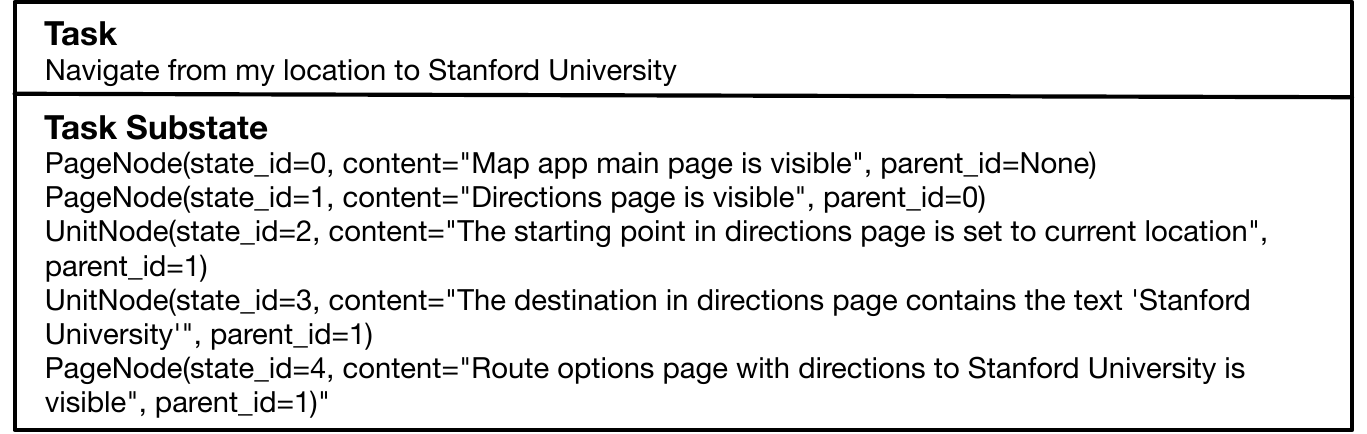}
\caption{Google Maps task with its substates}
\vspace{-0.5cm}  % 减少与下一个图片的间距
\label{fig:maps}
\end{figure*}

\begin{figure*}[htb]
\centering
\includegraphics[width=\textwidth]{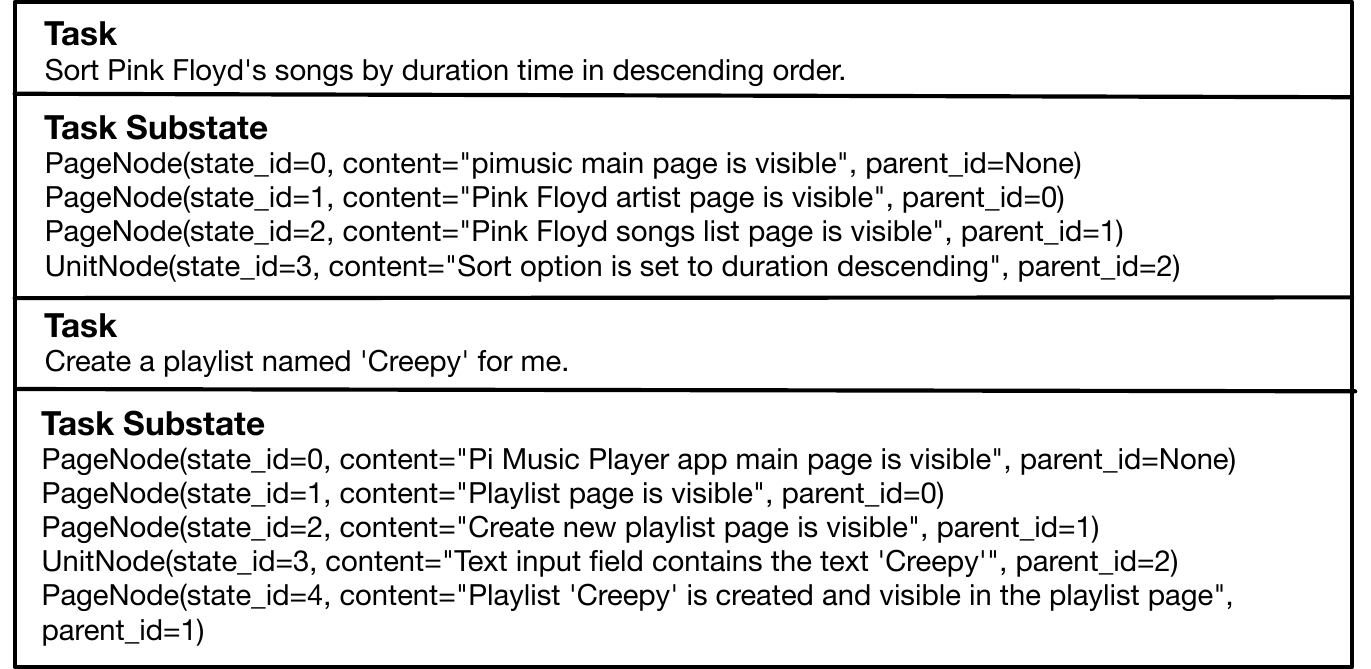}
\caption{Pi Music Player task with its substates}
\vspace{-0.5cm}  % 减少与下一个图片的间距
\label{fig:pimusic}
\end{figure*}

\begin{figure*}[htb]
\centering
\includegraphics[width=\textwidth]{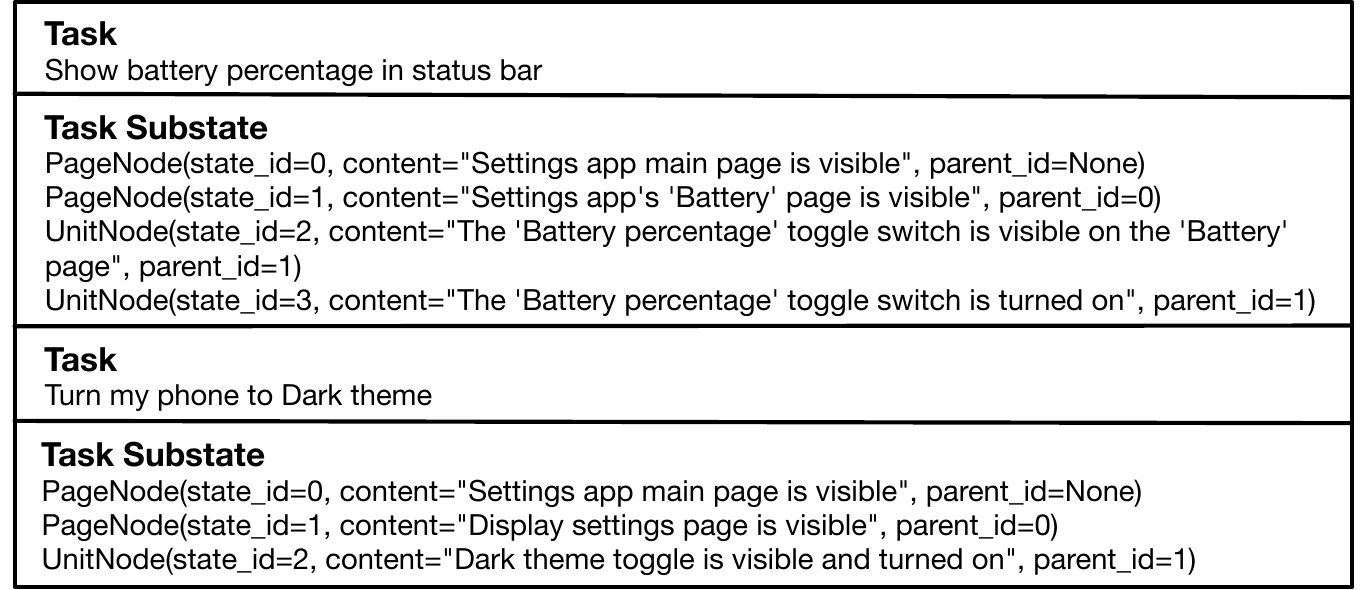}
\caption{Setting task with its substates}
\vspace{-0.4cm}  % 减少与下一个图片的间距
\label{fig:setting}
\end{figure*}

\begin{figure*}[htb]
\centering
\includegraphics[width=\textwidth]{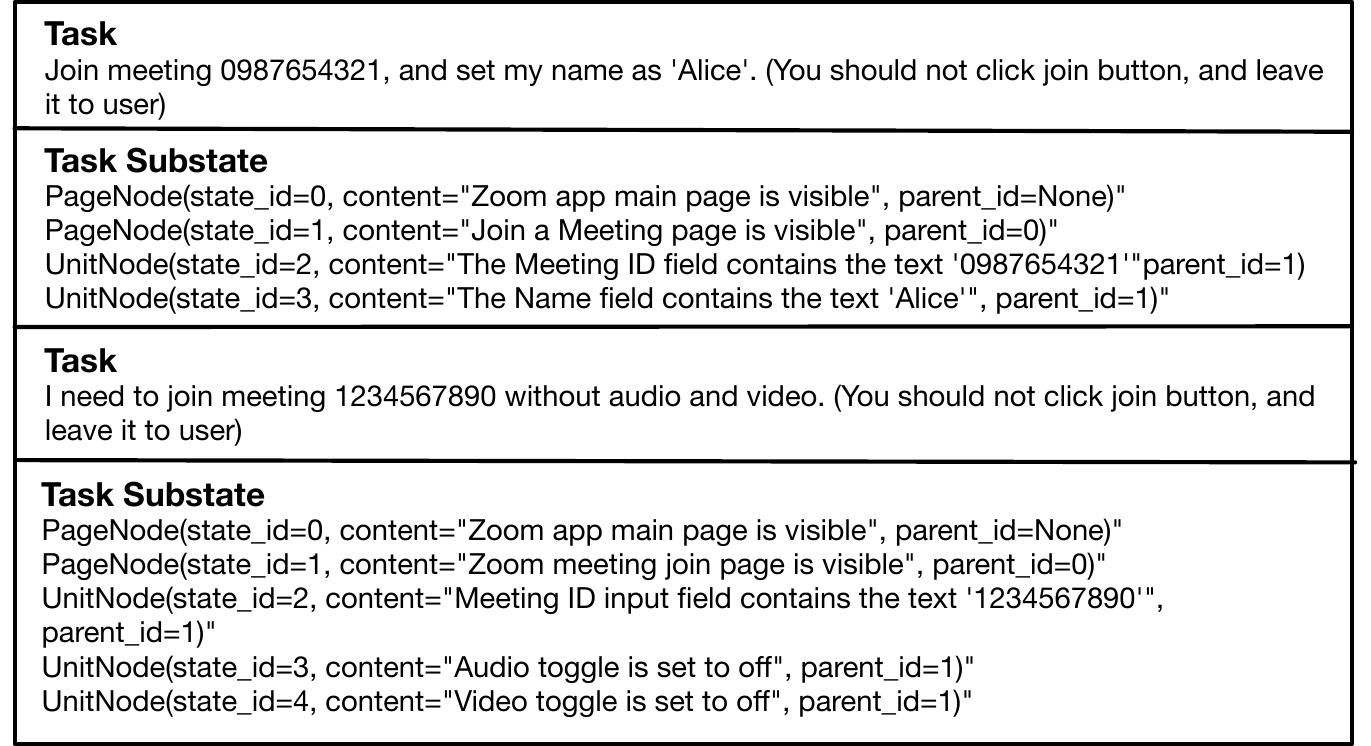}
\caption{Zoom task with its substates}
\label{fig:zoom}
\end{figure*}

\begin{figure*}[htb]
\centering
\includegraphics[width=0.8\textwidth]{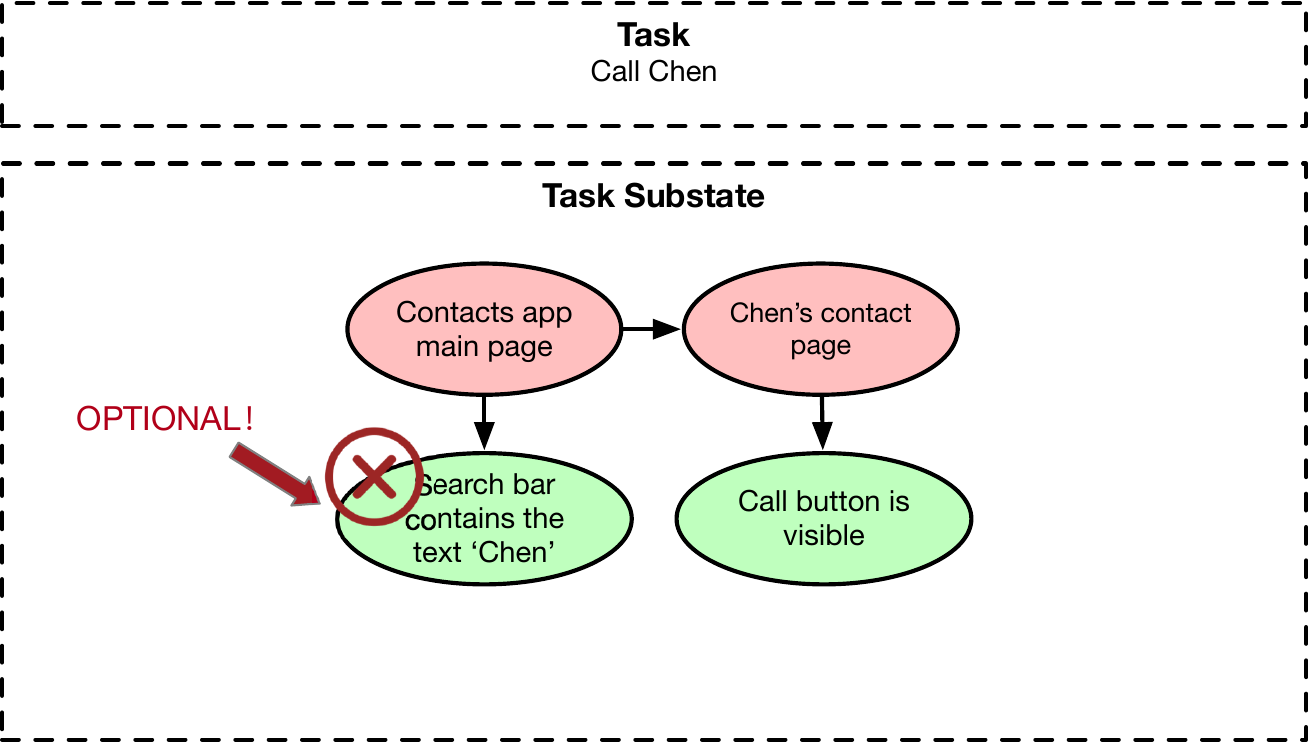}
\caption{Optional Case. Agent can open Chen's contact page without using search bar, so the substate "Search bar contains the text 'Chen'" is optional.}
\label{fig:optional-case}
\vspace{-0.2cm}  % 减少与下一个图片的间距
\end{figure*}

\begin{figure*}[htb]
\centering
\includegraphics[width=0.8\textwidth]{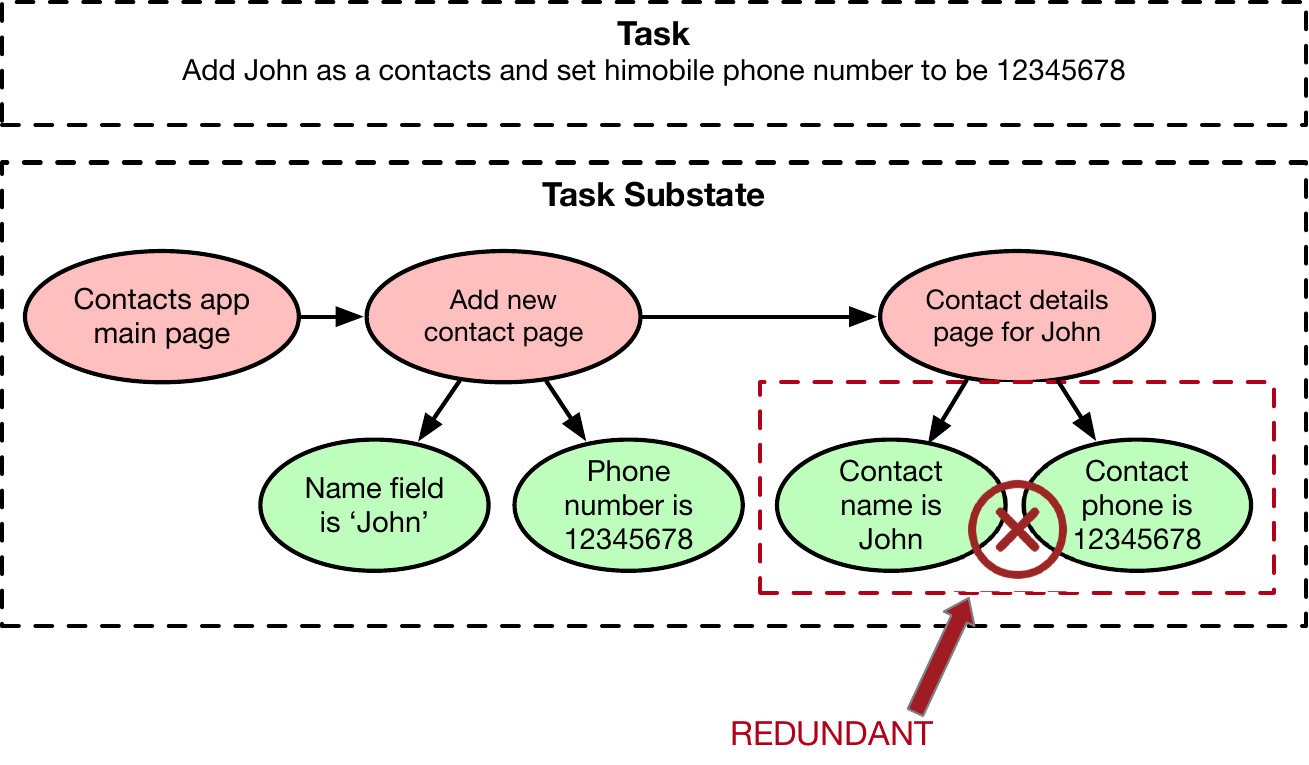}
\caption{Redundant Case. Child StateNode of 'Contact details page for John' is duplicate with child StateNodes of 'Add new contact Page'.}
\label{fig:redundant-case}
\vspace{-0.2cm}  % 减少与下一个图片的间距
\end{figure*}

\begin{figure*}[htb]
% \vspace{0.1cm}  % 减少与下一个图片的间距
\centering
\includegraphics[width=0.8\textwidth]{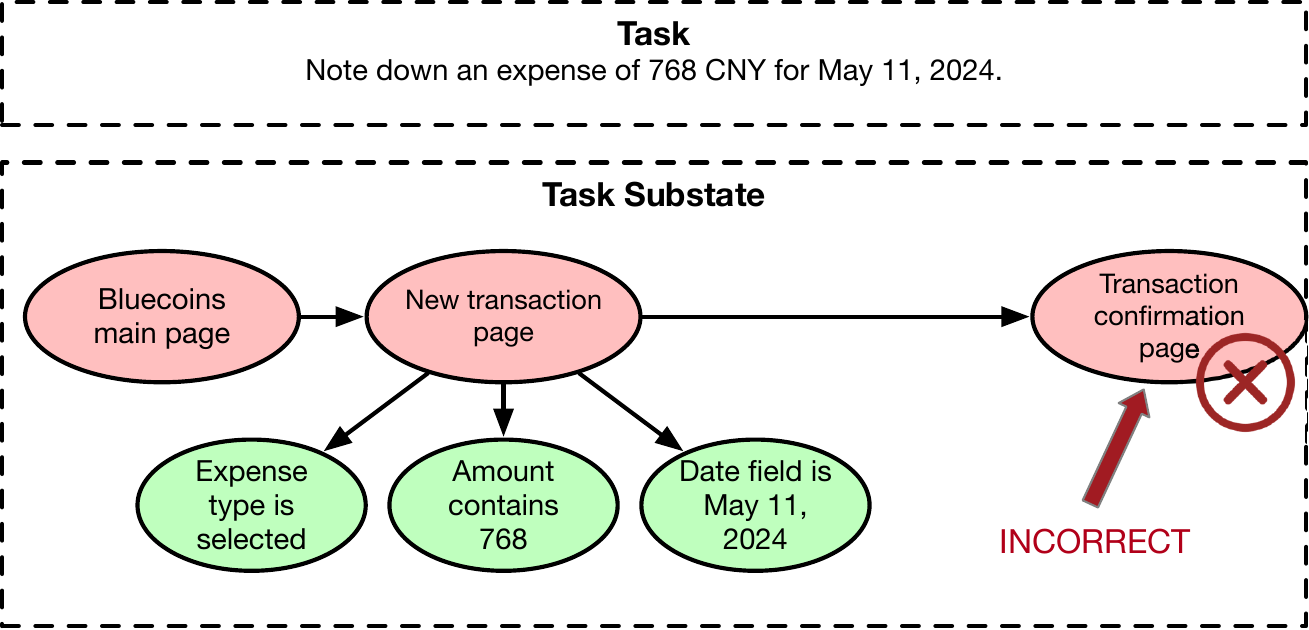}
\caption{Incorrect Case. The substate 'Transaction confirmation page' is non-existent in Bluecoins, so the substate is incorrect.}
\label{fig:incorrect-case}
\end{figure*}

%%%%%%%%%%%%%%%%%%%%%%%%%%%%%%%%%%%%%%%%%%%%%%%%%%%%%%%%%%%%%%%%%%%%%%%%%%%%%%%%
\end{document}